\documentclass[pdflatex,sn-nature]{sn-jnl}
\usepackage{graphicx}
\usepackage{multirow}
\usepackage{amsmath,amssymb,amsfonts}
\usepackage{amsthm}
\usepackage{mathrsfs}
\usepackage[title]{appendix}
\usepackage{xcolor}
\usepackage{textcomp}
\usepackage{manyfoot}
\usepackage{booktabs}
\usepackage{algorithm}
\usepackage{algorithmicx}
\usepackage{algpseudocode}
\usepackage{listings}
\usepackage{mathtools}
\usepackage{subcaption} 
\usepackage{geometry} 
\usepackage{lineno} 
\usepackage{float}
\usepackage{tabularx}
\theoremstyle{thmstyleone}

\theoremstyle{thmstyletwo}

\theoremstyle{thmstylethree}

\begin{document}
\title[Article Title]{\textbf{Spectral/Spatial Tensor Atomic Cluster Expansion with Universal Embeddings in Cartesian Space}}
\author[1,2]{\fnm{Zemin} \sur{Xu}}
\author*[1]{\fnm{Wenbo} \sur{Xie}}\email{\texttt{xiewb1@shanghaitech.edu.cn}}
\author[1,3]{\fnm{P.} \sur{Hu}}
\affil[1]{\orgdiv{School of Physical Science and Technology}, 
          \orgname{ShanghaiTech University}, 
          \city{Shanghai}, 
          \postcode{201210}, 
          \country{China}}
\affil[2]{\orgname{State Key Laboratory of Coordination Chemistry}, 
          \orgname{Key Laboratory of Mesoscopic Chemistry of MOE}, 
          \orgdiv{School of Chemistry and Chemical Engineering}, 
          \orgname{Nanjing University}, 
          \city{Nanjing}, 
          \postcode{210023}, 
          \country{China}}
\affil[3]{\orgdiv{School of Chemistry and Chemical Engineering}, 
          \orgname{The Queen's University of Belfast}, 
          \city{Belfast}, 
          \postcode{BT9 5AG}, 
          \country{UK}}
\abstract
{
    Equivariant atomistic machine learning models have largely been built on spherical-tensor representations, where explicit angular-momentum coupling introduces substantial complexity and systematic extensions beyond energies and forces remain challenging, often requires problem-specific architectural choices. Here we introduce the Tensor Atomic Cluster Expansion (TACE), which unifies scalar and tensorial modeling in Cartesian and space by decomposing local environments into irreducible Cartesian tensors (ICT) constructing a controlled many-body hierarchy with atomic cluster expansion (ACE). In addition to performing ACE in the frequency domain, we propose an efficient Clebsch-Gordan-free alternative in the spatial domain. TACE provides universal invariant (e.g., fidelity tags and charges) and equivariant (e.g., external electric fields and non-collinear magnetic moments) embeddings and predicted tensorial observables are handled on equal footing and enabling explicit control at inference. We demonstrate the accuracy, stability, and efficiency across finite molecules and extended materials, including in-domain and out-of-domain benchmarks, spectra, Hessian, external-field responses, charged systems, and multi-fidelity/head training. We further show its robustness on nonequilibrium/reactive datasets and controlled scaling when extending to large foundation model datasets.
}
\maketitle

\newpage
\section{Introduction}

    Atomistic simulations rely on accurate potential energy surfaces (PES) to describe structures and dynamics across molecules and condensed phases. While density functional theory (DFT) is the reliability reference, $\mathcal{O}(N^3)$ or higher scaling limits the accessibility of atomistic simulations, motivating machine learning interatomic potentials (MLIPs) as practical surrogates. MLIPs typically decompose the total energy into atomic site energies, which depend on the local environment of individual atoms. Early generations often relied on fixed hand-crafted representations and simple Cartesian neural networks that mapped atomic coordinates and species into invariant features. These invariant approaches achieved considerable success by ensuring translational, rotational, and permutational invariance~\citep{BPNN, MTP, REANN, SchNet, DeepMD}. However, they were fundamentally limited in accuracy and their ability to capture tensorial physical quantities. The advent of equivariant message passing neural networks marked a major breakthrough. By incorporating irreducible representations of the O(3) group via spherical tensor (ST), models such as NequIP, Allegro and MACE~\citep{NequIP, Allegro,MACE} introduced physics-informed inductive biases that significantly improved data efficiency and generalization. This shift revolutionized the field and made ST-based methods the mainstream choice. Nevertheless, spherical tensor formulations present intrinsic challenges. The use of Clebsch-Gordan (CG) coefficients to couple angular momenta incurs substantial computational cost. CG-free methods, such as the SO(2) tensor product~\citep{eSCN}, cannot be used for ACE, because these require one of the features to be fixed as a spherical harmonic. Moreover, ST are defined with respect to a fixed axis, commonly the $z$ axis, which introduces directional bias~\citep{ICT_book, ICTP}. 
    
    In parallel, Cartesian tensor approaches have experienced a resurgence as attractive alternatives to attempt to alleviate the burden of CG coefficients. Existing Cartesian approaches often rely on reducible Cartesian tensors, which carry redundant components~\citep{HotPP} and can complicate systematic control over symmetry constraints, particularly for higher-rank features and higher-body interactions~\citep{TensorNet}. Their accuracy has typically not matched that of models based on ST representations. CACE~\citep{CACE} encodes many body correlations explicitly but is limited in parameterization flexibility. CAMP~\citep{CAMP} enforced strict permutation symmetry but did not guarantee tracelessness. At the same time, irreducible tensors also exist in Cartesian space. Unlike ST, they do not privilege any orientation, and their tensor products are often simpler and more computationally efficient up to rank 4~\citep{ICTP}. ICTP~\citep{ICTP}, although pioneering the use of irreducible Cartesian tensors to achieve accuracies comparable to ST, requires explicit permutation of all indices in tensor products, and its implementation is limited to rank-3 tensors. Moreover, their representation cannot exploit lower-weight Cartesian tensors, thereby not fully realizing the potential of the Cartesian framework.
    
    On the other hand, although recent equivariant models have markedly advanced data efficiency and accuracy by learning from ab initio energies, forces, and stresses (E/F/S); can still yield limited fidelity in downstream physical property predictions despite low test errors when essential physical constraints are not included~\citep{eSEN} or when the models show poor extrapolation and robustness~\citep{Matbench}. Quantities readily available from electronic-structure calculations, such as charges, non-collinear magnetic moments, and field response tensors, are central to the physics but are often underutilized or not effectively integrated into current models.

    An ideal model should should therefore handle invariant and equivariant targets within a single framework. Co-supervising scalars such as energies, together with vectors and higher-rank tensors such as forces, stresses, polarizations, and polarizabilities, is important because they are not independent. Forces and stresses are derivatives of the same energy functional, and field-response tensors can encode additional information that can regularize the learned PES. At the same time, although direct-property prediction learning can be helpful in some special cases (e.g., direct forces in pretraining~\citep{eSEN}), robust and accurate prediction ultimately hinges on conservative training and physics-aware supervision. Likewise, magnetic materials are intrinsically 6N-dimensional rather than 3N (see SI Fig.~S1, so restricting supervision to E/F/S is insufficient for accurate predictions). More broadly, a model should accommodate both invariant variables (e.g., charge state or fidelity tags) and equivariant variables (e.g., external fields or non-collinear magnetic moments) within a single, shared, symmetry-consistent framework, enabling controlled extrapolation along physically meaningful directions that are otherwise entangled in the training.

    Here, we introduce the Tensor Atomic Cluster Expansion (TACE) to address these challenges by unifying scalar and tensorial modeling entirely within a single model in the Cartesian framework. By performing irreducible Cartesian tensor decomposition in Cartesian space, the number of arrays involved in tensor products is reduced to two and CG coefficients are no longer required for TACE in Cartesian space. Moreover, when implementing a spherical variant of TACE, the combined use of the SO(2) tensor product with spatial ACE eliminates the need for CG coefficients in spherical space. Through a basis transformation, our proposed spatial ACE can also be readily extended to Cartesian space.
    
    Building on this representation, TACE provides universal embeddings on both sides: (i) invariant embeddings for graph or node-level scalars (e.g., charges, fidelity tags, spin multiplicity) and (ii) equivariant embeddings for tensors (e.g., external fields, non-collinear magnetic moments). This design places predicted tensorial quantities and embedded tensorial inputs in the same symmetry-consistent feature space, enabling explicit control  during inference. A Latent Ewald Summation (LES)~\citep{LES1,LES2,LES3,LES4} plug-in further supplies a rigorous yet efficient route to long-range interactions within the same training loop. We perform comprehensive evaluations across in-domain and out-of-domain molecular sets, condensed-phase liquids, field-perturbed solids, charged materials, reactions systems in heterogeneous catalysis and universal MLIP benchmarks, to assess TACE's ability to integrate diverse physical information, both simultaneously and independently, to improve prediction accuracy. Across this breadth, TACE delivers accuracy on par with or surpassing that of leading methods, while providing a systematic and extensible paradigm for modeling the complex interplay between geometry, fields, and material properties.

\section{Results}

\begin{figure}[H]
    \centering
    \includegraphics[scale=0.76]{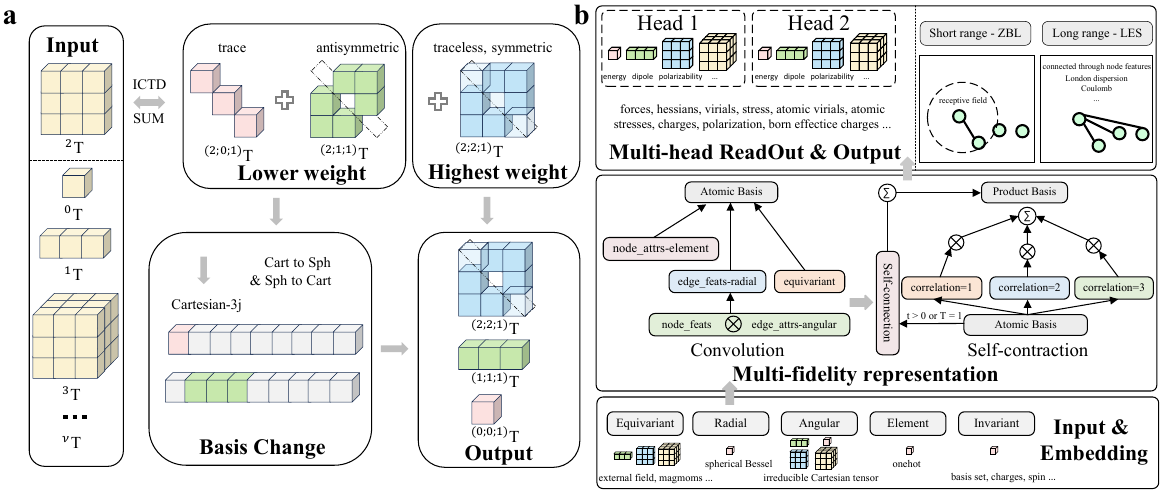}
    \caption{\textbf{Irreducible Cartesian tensor and the TACE architecture.}\label{fig:arch}
    \textbf{a} Taking the rank-2 case as an example, ICTD can extract the corresponding irreducible components with respect to a given weight. Through the basis transformation between Cartesian and spherical basis, the lower-weight parts of Cartesian tensors can be converted to ``$\mathrm{rank}=\mathrm{weight}"$.
    \textbf{b} Illustration of the overall data flow in TACE. The embeddings of invariants and equivariants are generic. In this work we demonstrate embeddings for different fidelity tags, charges, and external fields, but in principle other quantities are also supported. We also support the use of ZBL~\citep{ZBL} and Latent Ewald Summation~\citep{LES1, LES2, LES3, LES4} as plugins.}
\end{figure}

\subsection{Cartesian tensor}
\label{CTs}

    A generic Cartesian tensor of rank-$\nu$, denoted as $\prescript{\nu}{}{\mathbf{T}}$, is a multilinear object with $3^{\nu}$ components defined in a three-dimensional Euclidean space. The term ``\textbf{generic Cartesian tensor}" refers to one with no imposed symmetry or traceless constraints among its indices. These components transform according to specific representation rules under the action of the orthogonal group $\mathrm{O(3)}$, which includes both rotations and spatial inversions in three-dimensional space. In general, such tensors are reducible under $\mathrm{O(3)}$, meaning that they can be decomposed into a direct sum of irreducible components, each associated with a definite angular momentum character $\ell$. A generic Cartesian tensor $\prescript{\nu}{}{\mathbf{T}}$ can be expressed as a sum of ICT of the same rank but different weight $\ell$ (note that this ``weight" is not the ``learning weight" in machine learning):
    \begin{equation}
        \label{eqn:generic_tensor}
        \prescript{\nu}{}{\mathbf{T}} = \sum_{\ell, q} \prescript{{(\nu;\ell;q)}}{}{\mathbf{T}}
    \end{equation}
    where each $\prescript{{(\nu;\ell;q)}}{}{\mathbf{T}}$ is an irreducible component of rank-$\nu$, weight $l$ with $2\ell+1$ freedom, corresponding to a spherical tensor of degree $\ell$ and q denotes the label (since the multiplicity for given $\nu$ and $\ell$ may greater than 1).


\subsection{Irreducible Cartesian tensor decomposition}
\label{ICTD}
    As is well known, a generic rank-2 Cartesian tensor $\mathbf{T}$ can be decomposed into its irreducible components as follows (see also Fig.~\ref{fig:arch}a):
    \begin{equation}
        {\mathbf{T}}_{ij} = \Bigg( \underbrace{\tfrac{1}{2}({\mathbf{T}}_{ij} + {\mathbf{T}}_{ji}) - \tfrac{1}{3} \delta_{ij} {\mathbf{T}}_{kk}}_{\ell = 2}\Bigg)
        + \Bigg(\underbrace{\tfrac{1}{2}{(\mathbf{T}}_{ij} - {\mathbf{T}}_{ji})}_{\ell = 1}\Bigg)
        + \Bigg(\underbrace{\tfrac{1}{3} \delta_{ij} {\mathbf{T}}_{kk}}_{\ell = 0}\Bigg)
    \end{equation}
    This decomposition yields the rank-2 tensor's irreducible components corresponding to angular momenta $\ell=2$, $\ell=1$, and $\ell=0$. Generalizations to higher-rank tensors with $\nu = 3, 4, 5$ (numerical or analytical) were proposed in 1965, 1982, and 2024, respectively~\citep{rank3,rank4,rank5}. These ICT could be obtained via irreducible Cartesian tensor decomposition (ICTD) matrices. We denote the ICTD operater and its associated decomposition matrix as $\mathscr{T}$~\citep{ICTdecomposition}. Although it is possible to compute ICT with rank $\nu > 5$ using numerical methods, such approaches are not suitable for use in the forward pass of deep learning models, as they significantly degrade computational efficiency and as far as we know, there are no models utilizing numerical methods for this operation now. In fact, decomposition matrices were previously only available up to rank $\nu=5$, and their generation has traditionally been computationally expensive. However, thanks to the recent work of Shihao Shao~\citep{ICTdecomposition}, analytical ICTD matrices for $\nu > 5$ can now be efficiently constructed. The overall decomposition can be written as follows.
    \begin{equation}
        \operatorname{vec}\Big( \prescript{(\nu;\ell;q)}{}{\mathbf{T}} \Big)
        = \prescript{(\nu;\ell;q)}{}{C} \prescript{(\nu;\ell;q)}{}{C^T} \operatorname{vec}\Big( \prescript{\nu}{}{\mathbf{T}} \Big) =   \prescript{(\nu;\ell;q)}{}{\mathscr{T}} \operatorname{vec}\Big( \prescript{\nu}{}{\mathbf{T}} \Big)
    \end{equation}
   
\subsection{The TACE architecture}


    The feature learning of TACE begins with the construction of a graph in which each node is parameterized by learned ICT ${\mathbf{h}}_{i}$ that encode its chemical environment. For each edge, a one-particle basis $\boldsymbol{\phi}_{ij}$ is computed and to obtain permutation-invariant representations of semi-local environments, a summation pooling operation is applied to aggregate these local features, resulting in the so-called atomic basis ${\mathbf{A}}_{i}$. Building on this, higher-body-order features are systematically constructed through spectra or spatial ACE, yielding a product basis ${\mathbf{B}}_{i}$ that captures ($n + 1$)-body interactions. Overall, TACE follows the design framework of ACE~\citep{ACE, LinearACE, MACE, GRACE, BotNet}. Enabling parity allows the model to retain more intermediate irreducible components, but it has little effect on accuracy, as NequIP has shown~\citep{NequIP}. Therefore, in this work, we only implement model based on SO(3), although theoretically TACE is O(3)-equivariant. In addition to the aforementioned parts, some extra features have also been added, as shown in Fig.~\ref{fig:arch}b.

\subsubsection{Equivariant operation}
    Our architecture is primarily composed of two types of equivariant operations. The first type, tensor product/contracion maps two tensors $^{\nu_{1}}\mathbf{T}$ and $^{\nu_{2}}\mathbf{T}$ to a new tensor $^{\nu_3}\mathbf{T}$ whose rank $\nu_3 = \nu_1 + \nu_2 - 2k$ is determined by the the number of contracted indices $k$, where $0 \leq k \leq \min(\nu_1, \nu_2)$. When no indices are contracted, the operation is equivalent to tensor product. Below is the element-wise expression:
    \begin{align}
        \label{formula:tc}
        \prescript{{\nu_1 + \nu_2 - 2k}}{} 
        {\mathbf{T}_{a_1 \dots a_{\nu_1 - k} \, b_1 \dots b_{\nu_2 - k}}} 
        &= \frac{1}{\sqrt{3^k}} \cdot \sum_{i_1, \dots, i_{k}} \prescript{\nu_{1}}{}{\mathbf{T}_{a_1 \dots a_{\nu_1 - k} \, i_1 \dots i_{k}}} 
        \, \prescript{\nu_{2}}{}{\mathbf{T}_{i_1 \dots i_k \, b_1 \dots b_{\nu_2 - k}}}
    \end{align}
    We observe that for a given combination $(\nu_1, \nu_2, \nu_3)$, there can be multiple distinct ways to perform contractions, each corresponding to a different contraction path in Cartesian space. Drawing inspiration from e3nn~\citep{e3nn}, we define each such distinct contraction configuration as a path. Specifically, if $k$ indices are contracted, the number of possible paths can be computed without considering any symmetries. It equals the number of ways to select $k$ contracted indices from the first tensor and second tensor, multiplied by the number of permutations of these $k$ indices. Since Cartesian tensors correspond to reducible representations and therefore contain redundant information, Cartesian-based models~\citep{HotPP, CAMP} typically include only a single path for each combination and may directly embedding radial information to edge. In contrast, by employing ICTD, we have the flexibility to switch between irreducible and reducible Cartesian tensors. In the irreducible representation, the paths are often symmetric, allowing us to select only one of them. The second type of operation consists of linear combinations of tensors with identical rank $\nu$, resulting in new tensors of the same rank. Specifically, given $\tilde{c}$ input tensors $\prescript{\nu}{\tilde{c}}{\mathbf{T}}$, we can construct $c$ output tensors $\prescript{\nu}{c}{\mathbf{T}}$ of the same rank through a learnable linear transformation. 
    
\subsubsection{Node embedding}
    In the TACE framework, we denote node features by $\prescript{\nu_1}{c\,}{\mathbf{h}_{~i}^{(t)}}$, where $i$ refers to the atom index, $\nu_1 \in \mathrm{L_{max}}$ indicates the tensor rank, $p$ indexes the contributing paths, $c$ denotes the feature channel, and $t$ represents the layer index. The rank-0 node features are initialized by applying a (non)linear transformation to the one-hot encoding of the atomic species. Higher-rank node features are subsequently constructed via tensor contraction involving edge attributes. Node features will be iteratively updated through convolution across successive layers. The prevailing MLIPs predominantly adopt one-hot encoding to scalar features for reasons analogous to those in large language models (LLMs). Elements are not entirely independent, and those within the same period or group often exhibit similar properties. This correlation allows the learned element embeddings to be linearly reduced in dimensionality and visualized after training, thereby revealing underlying relationships. 

\subsubsection{Universal invariant embedding}
    Here, we proposed the incorporation of additional graph- or node-level invariant physical quantities such as charges and fidelity tags. For discrete element-like properties, we employ one-hot encoding followed by a linear layer, while for continuous properties we adopt an MLP-based embedding. All non-element invariant embeddings are concatenated and then projected into the channel dimension. We denote this combined representation as universal invariant features, which is then directly added to the node features. This embedding approach allows us to incorporate any invariant quantity that can reasonably influence the PES. By embedding these properties, the model gains additional descriptive flexibility for the PES, thereby improving its accuracy and transferability across different chemical environments.

    \textbf{Discrete embedding --- Multi-fidelity:} 
        In general, due to the computational cost of high accuracy computational level, MLIPs are often constructed at a reasonably chosen lower fidelity. However, a variety of approaches such as $\Delta$-learning, MFML, o-MFML, and MF$\Delta$ML~\citep{QeMFi-2} have been proposed to leverage abundant low fidelity data together with a limited amount of high fidelity data. This is commonly referred to as multi-fidelity training, where the term fidelity denotes the accuracy of the quantum chemistry method. These approaches typically aim to correct PES from low fidelity to high fidelity. Nevertheless, these methods are complex to apply, and we believe that their representation learning components are unable to achieve an optimal integration of low-fidelity and high-fidelity data. As an illustrative example, we consider the embedding of the same functional with different basis sets as a discrete variable to demonstrate the effectiveness of the universal invariant embedding. 

    \textbf{Continuous embedding --- Charges:} 
        For continuous variables, such as charges or the total charge, the embedding is performed in the same manner as described above and will not be elaborated further here.

\subsubsection{Universal equivariant embedding}
    Meanwhile, we propose a general scheme for embedding equivariant physical quantities within the ACE framework. This approach is universally applicable to other ACE-based models, allowing any physically meaningful equivariant features to be incorporated into the representation. 

    \textbf{External electric field:} The diverse responses of materials to external stimuli, ranging from linear to nonlinear and coupled effects, are crucial in determining the functional properties of a wide range of systems, including dielectrics, ferroelectrics, multiferroics, and piezoelectrics. Taking the external electric field as a representative example, we illustrate how such a field can be embedded into models based on ACE. Under universal equivariant embeddings, the $\mathbf{A}$-basis in Eqn.~\ref{formula:ab} is modulated by element- and channel-dependent weights $\mathrm{w_{c,z_i}}$, together with statistical information $\sigma$ about the field. This operation is applied after constructing the atomic basis and before forming the product basis. The atomic basis under an external electric field is then given by:
    \begin{equation}\label{formula:external_field}
        \underbrace{\prescript{1}{c}{\mathbf{A}}_{~i}^{(t)}}_{\mathrm{ext}}= \prescript{1}{c}{\mathbf{A}}_{~i}^{(t)} + \mathrm{w_{c,z_i}} \cdot \frac{\prescript{1}{}{\mathbf{E}_{\mathrm{ext}}}}{\sigma}
    \end{equation}
    \textbf{Non-collinear magnetic moments:} Magnetic materials present a particular challenge for MLIPs because, in addition to the 3N atomic positional degrees of freedom that define a conventional potential energy surface, they also carry 3N spin degrees of freedom associated with local magnetic moments. MagNet~\citep{MagNet} first utilized magnetic forces (i.e., the negative gradient of energy with respect to magnetic moments) as training labels. These advances now enable more thorough studies of magnetic systems. Leveraging the universal embedding framwork, we extend TACE to mTACE, which treats per-atom magnetic moments as explicit vector-valued features and learns their interactions on the same footing as structural degrees of freedom. In this case, the overall procedure remains essentially the same as in the presence of an external field, with the main distinction that the magnetic moments are treated for each atom. Although, due to the limited availability of public datasets and magnetic-related models, we do not present the benchmark results of mTACE here, we have already verified its effectiveness. Our method accounts for spin--lattice coupling, enabling the simultaneous optimization of magnetic moments and atomic coordinates, and it is already publicly available for use.

\subsubsection{Radial embedding}
    The spherical Bessel functions of the first kind, $j_n$, as proposed in DimeNet~\citep{DimeNet} are complete sets of orthogonal polynomials, and are therefore well-suited as radial basis functions. In this work, we adopt $j_0$ and construct the radial basis using its zeros. The corresponding expression is given by:
    \begin{equation}
        j_{0}^{n}(r_{ij}) = \sqrt{\frac{2}{c}}\frac{\sin(\frac{n\pi}{c}r_{ij})}{r_{ij}} f(r_{ij})
    \end{equation}
    The edge length $r_{ij} = \| \mathbf{r}_{ij} \|$ is first expanded using a set of radial basis functions and subsequently multiplied by an c2polynomial envelope cutoff function (or after radial MLP~\citep{mace-mh-1}) to ensure the smoothness of the PES. By tuning the hyperparameters of polynomial basis, we can effectively control the decay behavior of atomic contributions.~\citep{DimeNet}.
    \begin{equation}
    f(r_{ij}) =
    \begin{cases}
    1 - \frac{(p+1)(p+2)}{2} \left(\frac{r_{ij}}{c}\right)^p
    + p(p+2) \left(\frac{r_{ij}}{c}\right)^{p+1}
    - \frac{p(p+1)}{2} \left(\frac{r_{ij}}{c}\right)^{p+2}, & r_{ij} < c \\
    0, & r_{ij} \ge c
    \end{cases}
    \end{equation}
    It should be noted that the C2Polynomial cutoff is not continuous in the third derivative. Therefore, we propose the C3Polynomial cutoff:
    \begin{equation}
    f(r_{ij}) =
    \begin{cases}
    1 + a \left(\frac{r_{ij}}{c}\right)^p
    + b \left(\frac{r_{ij}}{c}\right)^{p+1}
    + c \left(\frac{r_{ij}}{c}\right)^{p+2}
    + d \left(\frac{r_{ij}}{c}\right)^{p+3}, & r_{ij} < c \\
    0, & r_{ij} \ge c
    \end{cases}
    \end{equation}

    with coefficients
    \begin{equation*}
    a = -\frac{(p+1)(p+2)(p+3)}{6}, \quad
    b = \frac{p(p+2)(p+3)}{2}, \quad
    c = -\frac{p(p+1)(p+3)}{2}, \quad
    d = \frac{p(p+1)(p+2)}{6}.
    \end{equation*}
    Other arbitrary differentiable envelope functions can generally also be used, although they may sometimes lead to a slight reduction in accuracy. This issue requires further investigation. The resulting radial bases are subsequently passed through MLP to generate path weights. We also implement $j_n$ of any order in the code. In our trials, using the first few orders did not result in a significant improvement or degradation of performance. Therefore, we adopt the safer low frequency $j_0$.

    \begin{equation}
        R_{p_1c\nu_1\nu_2\nu_3}^{(t)}(r_{ij})={\mathrm{MLP}}\Bigg(j_{0}^{n}(r_{ij})\Bigg)
    \end{equation}

\subsubsection{Cartesian harmonics}
    The angular information of an atomic environment are encoded in the normalized edge vector, $\hat{\mathbf{r_{ij}}} = \mathbf{r_{ij}} / r_{ij}$. In contrast to most of the SOTA models which rely on projections onto spherical harmonics , we adopt a cartesian representation. Specifically, we construct edge attributes $\prescript{(\nu_2;\nu_2;1)}{}{\mathbf{E}_{ij}}$, where $\nu_2 \in \ell_{max}$ as:
    \begin{align}
        \prescript{(\nu_2;\nu_2;1)}{}{\mathbf{E}_{ij}} &= \frac{(2\nu_2 - 1)!!}{\nu_2!}\cdot \prescript{(\nu_2;\nu_2;1)}{}{\mathscr{T}} \triangleright \Bigg(\underbrace{\hat{\mathbf{r_{ij}}} \boldsymbol{\otimes} \cdots \boldsymbol{\otimes} \hat{\mathbf{r_{ij}}} \boldsymbol{\otimes} \hat{\mathbf{r_{ij}}}}_{\nu_2 \text{ times}}\Bigg)
    \end{align}
    Different from cartesian models such as CAMP~\cite{CAMP} and HotPP~\citep{HotPP} using fully symmetric tensor directly, we futher reduce redundant traces, utilizing its highest-weight irreducible components, fully symmetric, traceless Cartesian tensors which serve as the Cartesian analogue of spherical harmonics. Edge attributes with higher rank can capture more fine-grained directional information. In practice, we typically set $\ell_{\mathrm{max}} = 3$, consistent with common choices in ST models. The constant here is used to ensure that $\prescript{(\nu_2;\nu_2;1)}{}{\mathbf{E}_{ij}}$ can contract to $\hat{\mathbf{r_{ij}}}$~\citep{Angular}.

\subsubsection{Atomic basis}
    We begin by constructing the one particle basis $\prescript{{\nu_3}}{p,c}{\boldsymbol{\phi}}_{ij}^{(t)}$ where ${\nu_3} \in \mathrm{\ell_{max}}$ in Eqn.~\ref{formula:opb} via tensor contraction between node features and edge attributes. Each path is modulated by a learnable radial weight. The particle basis indeed captures pairwise ($2$-body) interactions at the first layer and higher-body interactions at the rest. The $\mathbf{A}$-basis in Eqn.~\ref{formula:ab} is then formed by aggregating the particle basis over neighboring atoms, followed by decomposition to get highest-weight irreducible components. It should be noted that the normalization constant $\mathcal{N}(i)$ can be dynamically determined based on node features, node attributes or edge features, or it can be set according to the average number of neighboring atoms.
    \begin{equation}\label{formula:opb}
        \prescript{\nu_3}{p_1,c~}{\boldsymbol{\phi}}_{ij}^{(t)} =  R_{~p_1c\nu_1\nu_2\nu_3}^{(t)}(r_{ij}) \cdot \Bigg(\prescript{\nu_1}{c\,}{\mathbf{h}_{~i}^{(t)}} \otimes_R^{(\nu_1,\nu_2,{\nu_3},p_1)} \prescript{(\nu_2;\nu_2;1)}{}{\mathbf{E}_{ij}}\Bigg)
    \end{equation}
    \begin{equation}\label{formula:ab}
        \prescript{\nu_3}{c\,}{\mathbf{A}}_{i}^{(t)}=\frac{1}{{\mathcal{N}(i)}} \cdot \prescript{(\nu_3;\nu_3;1)}{}{\mathscr{T}} \triangleright \Bigg(\sum_{p_1,\tilde{c}}W_{p_1\tilde{c}\nu_3c}^{(t)}\sum_{j\in\mathcal{N}(i)}\prescript{{\nu_3}}{p_1,c}{\boldsymbol{\phi}}_{ij}^{(t)}\Bigg)
    \end{equation}

\subsubsection{Product basis}

    \textbf{Cartesian spectral ACE.}
        Once the atomic basis is obtained, $\mathbf{B}$-basis in Eqn.~\ref{formula:pb} can be constructed via self-contraction. The n-correlation $\prescript{\nu_4}{}{\mathbf{B}_{i,n}}$ where $ \nu_4 \in \mathrm{L_{max}}$ is obtained via n times contracton with $\mathbf{A}$-basis itself. Note that after every tensor contracion, we should apply an decomposition matrix to obtain highest weight.
        \begin{equation}\label{formula:pb}
            \prescript{\nu_4}{p_2c\,}{\mathbf{B}}_{i,n}^{(t)} = \underbrace{\bigotimes_{\xi=1}^{n} \prescript{{(\nu_3;\xi)}}{c\,}{\mathbf{A}}_{~i}^{(t)}}_{\text{with (n-1) times } \mathscr{T}} 
        \end{equation} \\
    \textbf{Spherical spatial ACE.}
        Since the learned node features are SO(3) equivariant, they are essentially the coefficients of functions on the sphere. In order to preserve SO(3) equivariance, we cannot arbitrarily apply pointwise operations. At the same time, the coupling of the product basis in recursive high-order correlation calculations is computationally expensive and also involves CG coefficients, which severely limits the spectral operations that can be performed. Here, we propose a method to perform atomic cluster expansion in spatial domain. 
    
        The atomic basis $\mathrm{A_{i}}$ in spatial domain is defined as:
        \begin{equation}
            \mathrm{A_{i\tilde{b}bac}} = \sum_{m} \mathrm{SF_{bam}^{-1} \mathbf{A_{i\tilde{b}mc}}} 
        \end{equation}
        
        The product basis $\mathrm{B_{i, n}}$ in frequency domain is defined as:
        \begin{align}
        \mathbf{B_{i\tilde{b}mc}} &= \sum_{ba} \mathrm{SF_{bam}} \Bigl( 1.0 \cdot \mathrm{A_{i\tilde{b}bac}^1} \,\|\, \sigma_2 \cdot \mathrm{A_{i\tilde{b}bac}^2} \,\dots\, \|\, \sigma_n \cdot \mathrm{A_{i\tilde{b}bac}^n} \Bigr)
        \end{align}
        where $i$ represents the atom index, $n$ represents the feature dimension, which is equal to $\sum_\ell 2\ell+1$, $\tilde{b}$ represents the batch dimension, $a$ represents the longitude resolution, $b$ represents the latitude resolution, $c$ represents the channel and $SF^{-1}$ denotes the inverse spherical Fourier transform, $SF$ denotes the spherical Fourier transform, $\sigma$ represents a learnable scalar scale and $||$ denotes concatenation along the last dimension. \\
    \textbf{Cartesian spatial ACE.}
        Since the transformation between ICT and ST differs only by a basis transformation via the path matrix, we only need to define the Cartesian spherical Fourier transform and its inverse as:
        \begin{align}
        \mathrm{CF_{ban}} &= \Bigl(\sum_{m_0} \mathrm{F_{bam_0} \, \prescript{(0;0;1)}{}{C_{n_0 m_0}}}\Bigr)
        || \Bigl(\sum_{m_1} \mathrm{F_{bam_1} \, \prescript{(1;1;1)}{}{C_{n_1 m_1}}}\Bigr)
        || ...
        || \Bigl(\sum_{m_\ell} \mathrm{F_{bam_\ell} \, \prescript{(l;l;1)}{}{C_{n_\ell m_\ell}}}\Bigr) \\
       \mathrm{CF_{ban}^{-1}} &= \Bigl(\sum_{m_0} \mathrm{F_{bam_0}^{-1} \, \prescript{(0;0;1)}{}{C_{n_0 m_0}}}\Bigr)
        || \Bigl(\sum_{m_1} \mathrm{F_{bam_1}^{-1} \, \prescript{(1;1;1)}{}{C_{n_1 m_1}}}\Bigr)
        || ...
        || \Bigl(\sum_{m_\ell} \mathrm{F_{bam_\ell}^{-1} \, \prescript{(l;l;1)}{}{C_{n_\ell m_\ell}}}\Bigr)
        \end{align}
    where $C$ represents the path matrix, $m_l$ equals to $2\ell+1$,  $n_l$ equals to $3^{\ell}$, $CF^{-1}$ denotes the inverse Cartesian Fourier transform and $CF$ denotes the Cartesian Fourier transform.

\subsubsection{Message passing}
    In TACE, the combination of equivariance, path and higher-body-order interactions allows the model to achieve the SOTA accuracy with only two layers (one message passing with T=2), outperforming Cartesian models such as CMAP~\citep{CAMP} and HotPP~\citep{HotPP} typically with T $>2$.
    \begin{align}\label{formula:coupled}
        \prescript{\nu_4}{c\,}{\mathbf{h}_{i}^{(t+1)}} =
        \begin{cases}
            \sum_{np_2\tilde{c}c}W_{
            np_2\tilde{c}z_i\nu_4}^{(t)}\prescript{\nu_4}{p_2c\,}{\mathbf{B}}_{i,n}^{(t)} & \text{if } (\mathrm{t=0}) \\
            \sum_{np_2\tilde{c}c}W_{
            np_2\tilde{c}z_i\nu_4}^{(t)}\prescript{\nu_4}{p_2c\,}{\mathbf{B}}_{i,n}^{(t)} + \sum_{\tilde{c}}W_{z_i \nu_4\tilde{c}c}^{(t)}\prescript{\nu_4}{c\,}{\mathbf{A}}_{i}^{(t)} & \text{if } (\mathrm{t>0}) 
        \end{cases}
    \end{align}
    The node features at layer $(t+1)$ are computed according to Eqn.~\ref{formula:coupled}. It is important to note that there are many possible choices for residual connections, for example, connecting $\mathbf{A}$ to $\mathbf{B}$ within the current layer, connecting $\mathbf{B}$ from the previous layer to $\mathbf{A}$ in the current layer, or patterns such as $\mathbf{BAB}$, etc.~\citep{Residual}. To enhance the expressive power of the model, we employ coupled features. The TACE framework further provides flexibility to choose between coupled features and element-dependent weights.
    
\subsubsection{Readout}
    The model predicts short range site energies as a learnable function of the rotationally invariant components of the node features. To preserve the body-order structure inherent in the model. The readout function is kept linear across all but the final layer. In practice, the readout function combines linear neural networks and MLPs, the linear term retains contributions consistent with a body-ordered expansion, while the MLPs capture higher-order interactions from the truncated series. Forces on atoms are obtained via differentiation of the total energy using automatic differentiation. Naturally, the site energy is scaled and shifted, with the specific approach depending on the user's choice.
    \begin{align}
        \mathcal{R}^{(t)}\left(\prescript{0}{c}{h_{~i}^{(t)}}\right) &=
        \begin{cases}
            \sum_c W_c^{(t)} \prescript{0}{c}{h_{~i}^{(t)}} & \text{if } \mathrm{t < T-1}, \\
            \mathrm{MLP}\left(\prescript{0}{c}{h_{~i}^{(T)}}\right) & \text{if } \mathrm{t = T-1}
        \end{cases} \\
        \mathrm{E_{i}^{sr}} = \mathrm{E_{i}^{ele}} + \mathrm{E_{i}^{uie}} + \sum_{t=0}^{T-1} \mathrm{E}_{i}^{(t)} &= \mathrm{E_{i}^{ele}} + \mathrm{E_{i}^{uie}} + \sum_{t=0}^{T-1} \mathcal{R}^{(t)}\bigg(\prescript{0}{c}{h_{~i}^{(t)}}\bigg)
    \end{align}
    For tensors that are directly predicted, such as the dipole moment, polarizability, direct forces, direct stress, and so on, we only need to add a readout corresponding to the appropriate rank. Taking the polarizability as an example, it is constructed from a combination of rank-0 and rank-2 tensors to yield a fully symmetric tensor. Finally, we emphasize that the energy prediction employs the isolated atomic energy either obtained through least squares fitting or provided directly by the user from DFT calculations as the baseline. In addition, when universal invariant features are available, we can further obtain an optional baseline energy through universal invariant features.

\subsubsection{Charges equilibration}
    In principle, for any node-level rank-0 tensor that must be conserved, the surplus part of the prediction can be uniformly redistributed among atoms. As an illustrative example, we consider the case of atomic charges. While such uniform redistribution is possible, a more physically constrained choice is the Charge Equilibration approach~\citep{EEM,4G-HDNNPs}. To this end, we introduce two additional rank-0 readout MLP to predict the electronegativity $\chi$ and hardness $\eta~\mathrm{(> 0)}$  or $\frac{1}{\eta}$ for each atom. The atomic charges are then obtained by enforcing charge conservation through the following expression:

    \begin{equation}\label{formula:charges}
        \mathrm{Q_i = \frac{\frac{Q_{tot} + \sum_i\frac{\chi_i}{\eta_i}}{\sum_i\frac{1}{\eta_i}} - \chi_i}{\eta_i}}
    \end{equation}

\subsubsection{Latent Ewald Summation}
    For strict-local and semi-local models based on the short-range approximation, although they are highly effective for most systems, they often exhibit unphysical behavior in systems where long-range interactions are significant, due to neglecting atoms outside the receptive field. Here, we employ the Latent Ewald Summation (LES) as a long-range-correction-drop-in module. For the sake of completeness in model formulation, we briefly outline the principle of LES and its implementation within TACE. Additional details on the implementation and underlying physics can be found in refs.~\citep{LES1, LES2, LES3, LES4}. 
    The long-range correction energy for extend system can be fourmulated as:
    \begin{equation}
        \mathrm{S(\mathbf{k})=\sum_{i=1}^Nq_i^\mathrm{les}e^{i\mathbf{k}\cdot\mathbf{r}_i}}
    \end{equation}
    \begin{equation}
        \mathrm{E^{lr}=\frac{1}{2\varepsilon_0V}\sum_{0<k<k_c}\frac{1}{k^2}e^{-\sigma^2k^2/2}|S(\mathbf{k})|^2}
    \end{equation}
    where $\varepsilon_{0}$ is the vacuum permittivity, $k$ is the wave vector determined by the cell of the system, $k_c$ is the $\mathbf{k}$-point cutoff (typically chosen to be $\pi$). $V$ is the cell volume, and $\sigma$ is a smearing factor (typically chosen to be $1\,\text{\AA}$).
    For finite systems, the long-range energy is computed using the pairwise direct sum:
    \begin{equation}
        \mathrm{E^{lr}=\frac{1}{2}\frac{1}{4\pi\varepsilon_0}\sum_{i=1}^N\sum_{j=1}^N[1-\varphi(r_{ij})]\frac{q_i^{les}q_j^{les}}{r_{ij}}}
    \end{equation}
    where the complementary error function $ \varphi(r) = \mathrm{erfc}(\frac{r}{\sqrt{2}\sigma})$. It should be noted that, depending on the functional form, LES can simultaneously describe both electrostatic and dispersion interactions. Hereafter, the model employing LES as a plugin is referred to as TACE-LES.

\begin{table}[]
    \tiny
    \centering
    \caption{RMSE for energy (E, meV) and forces (F, meV/Å) for the 3BPA dataset. The best-performing results are both \textbf{bolded} and \underline{underlined}.}
    \label{tab:3bpa_results}
    \begin{tabularx}{\textwidth}{ccccccccccc}
        \toprule
        &   
        & \textbf{ACE}\citep{ACE} 
        & \textbf{CACE}\citep{CACE} 
        & \textbf{Allegro}\citep{Allegro} 
        & \textbf{NequIP}\citep{NequIP}  
        & \textbf{MACE}\citep{MACE} 
        & $\mathbf{TACE}_{\mathrm{spa}}^{\mathrm{sph}}$ 
        & $\mathbf{TACE}_{\mathrm{spa}}^{\mathrm{cart}}$ 
        & $\mathbf{TACE}_{\mathrm{spe}}^{\mathrm{cart}}$ 
        & $\mathbf{TACE}_{\mathrm{spe}}^{\mathrm{cart}}$  \\
        \midrule
        300K & E & 7.1   & 6.3   & 3.8  & 3.3 & 3.0  
             & 4.0 & 3.20 & 3.5 & \underline{\textbf{2.69}}  \\
             & F & 27.1  & 21.4  & 13.0 & 10.8  & 8.8  
             & 10.2 & 10.0 & 10.2 & \underline{\textbf{8.68}} \\
        \addlinespace
        600K & E & 24.0  & 18.0  & 12.1 & 11.2  & 9.7   
             & 11.7 & 12.9 & 11.6 & \underline{\textbf{8.10}} \\
             & F & 64.3  & 45.2  & 29.2 & 26.4  & 21.8 
             & 24.1 & 23.80 & 23.4 & \underline{\textbf{19.88}}\\
        \addlinespace
        1200K & E & 85.3  & 58.0  & 42.6 & 38.5  & 29.8  
               & 30.4 & 30.6 & 29.7 & \underline{\textbf{23.42}} \\
              & F  & 187.0 & 139.6 & 83.0 & 76.2 & 62.0 
              & 63.8 & 65.8 & 60.5 & \underline{\textbf{54.26}} \\
        \addlinespace
        dihedral & E  & -    & -    & -   & 23.2  & 7.8  
                 & 14.4 & 11.4 & \underline{\textbf{7.6}} & 11.00 \\
                 & F  & -    & -    & -   & 23.1  & \underline{\textbf{16.5}} 
                 & 18.8 & 18.3 & 18.9 & \underline{\textbf{16.45}} \\
        \bottomrule
    \end{tabularx}
    \footnotetext{All models were trained using dataset at $\mathrm{T} = 300 \, \mathrm{K}$. Results for all models except TACE are taken from ref.~\citep{CACE}. From left to right, the first three TACE models use correlation order 2, while the last model uses correlation order 3. ``spa'' denotes the spatial domain, ``spe'' denotes the frequency domain, ``sph'' indicates the spherical representation, and ``cart'' indicates the Cartesian representation.}

\end{table}

\subsection{Out-of-Domain --- 3BPA}
    \label{3bpa}
    We evaluate the extrapolation capabilities of TACE on the 3-(benzyloxy)pyridin-2-amine (3BPA) dataset, originally introduced in ref.~\citep{LinearACE}, which comprises a flexible, drug-like molecule with three rotatable single bonds. The training set consists of 500 geometries sampled from a 300K MD trajectory, which were subsequently reevaluated at the DFT level using the $\omega$B97X exchange-correlation functional and the 6-31G(d) basis set.
    To assess both in-domain and out-of-domain generalization, three test sets are constructed using MD trajectories at 300K, 600K, and 1200K, respectively. An additional fourth test set consists of dihedral scan geometries, where two dihedral angles are held fixed and the third is systematically varied from 0 to 360 degrees. This results in dihedral slices that sample regions of the PES significantly outside the domain represented by the training data.

    Across all splits, TACE attains force and energy RMSE competitive with or better than leading equivariant models, while using only 64 channels compared to 256 in a comparable MACE baseline (Table~\ref{tab:3bpa_results}.)  Moreover, the 3BPA dataset is employed in our study to investigate the differences between coupled and uncoupled feature representations in Eqn.~\ref{formula:coupled}. We observe that coupled features consistently achieve higher accuracy on force predictions across all test sets. Nonetheless, this result indicates that models may achieve higher accuracy with fewer channels when using coupled channel features. 

\begin{figure}[]
    \centering
    \includegraphics[scale=0.34]{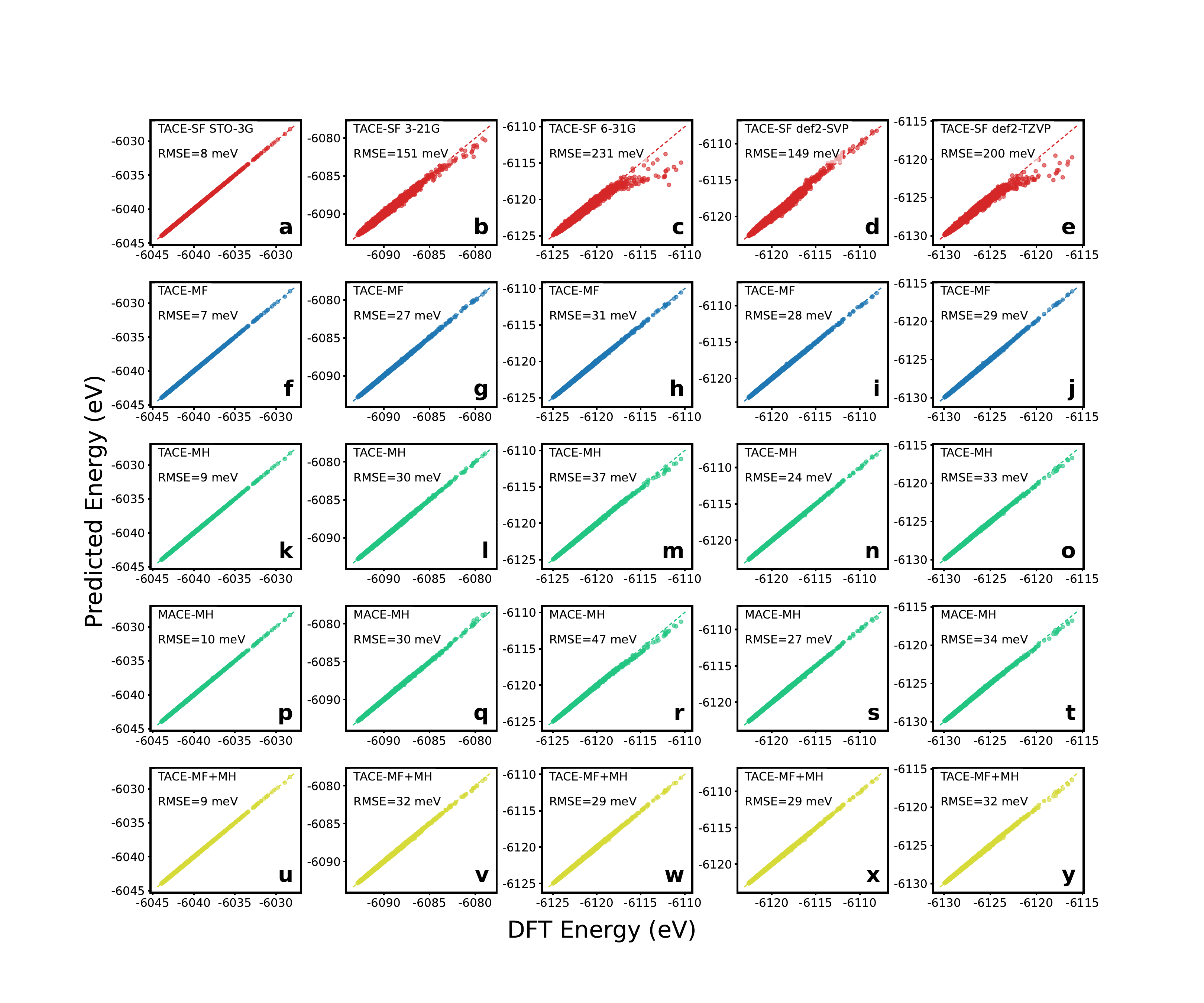}
    \caption{\textbf{Comparison of Different Mixed Precision Training Approaches.} From left to right, the basis sets are STO-3G, 3-21G, 6-31G, def2-SVP, and def2-TZVP. The test set is the same for each column. \textbf{a-e} show TACE using single-fidelity training. \textbf{f-j} show TACE using multi-fidelity training. \textbf{k-o} show TACE using multi-head training, \textbf{p-t} show MACE using multi-head training. \textbf{u-y} show TACE using both multi-fidelity and multi-head schemes.}
\label{fig:urea}
\end{figure}

\subsection{Mixed Precision Training --- Urea} 

    Training on datasets with mixed precision has emerged as an important strategy to reduce the cost of generating high-accuracy reference data. Here, we treat the fidelity tag as an additional discrete label using the universal invariant embedding framework and train TACE on datasets of different electronic-structure fidelities within a single unified model. This strategy ,which we refer as multi-fidelity training, differs conceptually from standard multi-head training, implemented in MACE~\citep{MACE, MACE-uMLIP}. 
    In multi-head training, separate output heads are attached for each fidelity level while sharing a common representation; in multi-fidelity training, the fidelity label is embodied directly into the representation itself, so that the internal features become level-aware. Both strategies can coexist, but multi-fidelity training offers a more natural route to cross-fidelity generalization.

    We evaluate this idea on the  QeMFi~\citep{QeMFi-1} dataset to compare the effects of single-fidelity (SF), multi-fidelity (MF), multi-head (MH), and their combined approaches on predictive performance. The dataset contains 15,000 molecular configurations for each of five levels, computed using TD-DFT with the CAM-B3LYP functional. The basis sets, in order of increasing accuracy, are: STO-3G, 3-21G, 6-31G, def2-SVP, and def2-TZVP.  A large low-fidelity dataset supplemented with a small amount of high-fidelity data: STO-3G serves as the baseline with a train:validation:test split of 41:9:50, while the other levels use a split of 1:9:90, so that only a very small amount of the total training data comes from the higher levels. Single-fidelity models are trained using datasets from only one basis set, whereas other models are trained on a combined dataset incorporating all five fidelity levels. 
    As shown in Fig.~\ref{fig:urea}, the red region indicates that, for the high-fidelity dataset, the model trained solely on limited high-fidelity data fails to generalize to unseen configurations. In contrast, both multi-fidelity and multi-head training greatly reduce these errors as high-fidelity supervision is now supported by abundant low-fidelity structures. Multi-fidelity TACE achieves the lowest errors across high-energy configurations, outperforming both the multi-head variant of TACE and the multi-head MACE baseline under matched architectures and hyperparameters. Although the yellow region in Fig.~\ref{fig:urea} does not show a significant advantage from combining multi-head and multi-fidelity schemes, this combination still provides a more flexible modeling option. It is also worth noting that the introduction of high-fidelity data may slightly degrade the accuracy on low-fidelity datasets, but the gain at the high-fidelity levels is substantial. Overall, these experiments demonstrate that universal embeddings let TACE act as a cross-fidelity learner.

\begin{table}[h]
    \tiny
    \centering
    \caption{
        RMSE for energy per atom (E, meV/atom) and forces (F, meV/Å) for the 32-water system. 
        The best-performing results are highlighted in both \textbf{bold} and \underline{underline}.
    }
    \label{tab:h2o_results}
    \begin{tabularx}{\textwidth}{cccccccccccc}
        \toprule
        & \multicolumn{4}{c}{Invariant Models} & \multicolumn{3}{c}{Equivariant Models} & \multicolumn{2}{c}{MACE}\citep{MACE} & \multicolumn{2}{c}{TACE} \\
        \cmidrule(lr){2-5} \cmidrule(lr){6-8} \cmidrule(lr){9-10} \cmidrule(lr){11-12}
        & \textbf{BPNN}\citep{BPNN} & \textbf{DeePMD}\citep{DP} & \textbf{ACE}\citep{LinearACE} & \textbf{REANN}\citep{REANN} & \textbf{NequIP}\citep{NequIP} & \textbf{CACE}\citep{CACE} & \textbf{CAMP}\citep{CAMP} & \textbf{64-0} & \textbf{192-2} & \textbf{64-0} & \textbf{64-2} \\
        \midrule
        E & 2.3 & 2.1 & 1.7 & 0.8 & 0.9 & \underline{\textbf{0.6}} & \underline{\textbf{0.59}} & \underline{\textbf{0.6}} & \underline{\textbf{0.6}} & \underline{\textbf{0.58}} & 0.72 \\
        F & 120 & 92 & 99 & 53 & 45 & 47 & 34 & 37.1 & 36.2 & 33.16 & \underline{\textbf{30.31}} \\
        \bottomrule
    \end{tabularx}
    \footnotetext{Results for all models except TACE are taken from refs.~\citep{CAMP, MACE-application}. Here, 64-0 denotes $c=64$ and $\mathrm{L_{max}}=0$.}
\end{table}
\begin{figure}[h]
    \centering
    \includegraphics[scale=0.35]{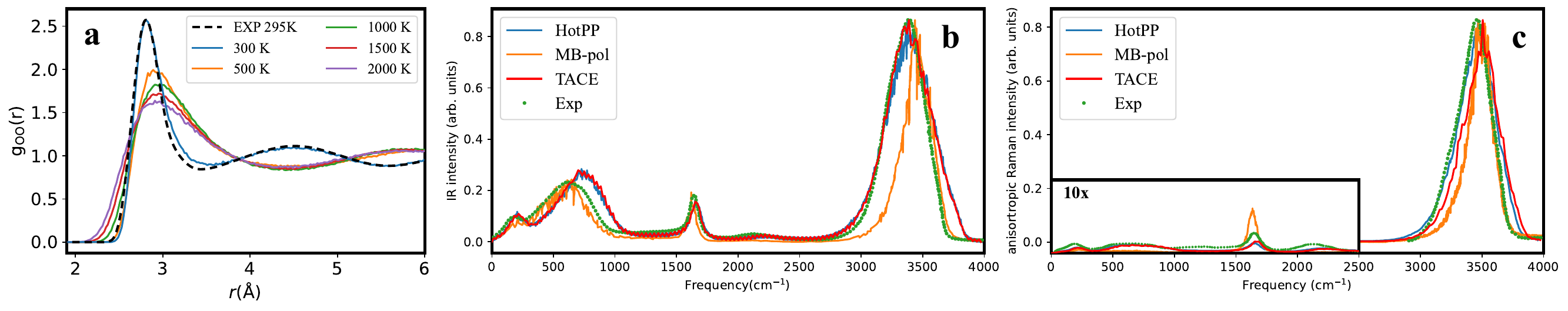}
    \caption{\textbf{Water simulation outcomes with $\mathbf{L_{max}=0}$.} \textbf{a} O-O RDFs at different temperatures from NVT MD simulations and X-ray diffraction. \textbf{b} IR spectra. \textbf{c} Raman spectra.}
    \label{fig:h2o}
\end{figure}

\begin{table}[ht]
    \centering
    \caption{Relative RMSE, defined as the RMSE normalized by the intrinsic standard deviation of the validation set (RRMSE, \%) for dipole moment (D) and polarizability ($\alpha$) across different water systems. For each training set size, the best-performing results are highlighted in both \textbf{bold} and \underline{underline}.}
    \label{tab:dipole_and_polar}
    \begin{tabular}{cccccccc}
    \toprule
        &  & \multicolumn{4}{c}{$N_{\text{train}}=500$} & \multicolumn{2}{c}{$N_{\text{train}}=700$} \\
        \cmidrule(lr){3-6} \cmidrule(lr){7-8}
        System &  & SA-GPR\citep{SA-GPR} & REANN\citep{REANN} & TNEP\citep{TNEP} & TACE & HotPP\citep{HotPP} & TACE \\
        \midrule
        monomer & D & 0.11 & 0.05 & 0.069 & \underline{\textbf{0.008}} & 0.02 & \underline{\textbf{0.006}} \\
        & $\alpha$ & 0.02/0.12 &  \underline{\textbf{0.06}} & 5.89/1.22 & 0.086 & 0.05 & \underline{\textbf{0.042}} \\
        \addlinespace
        dimer   & D & 5.3 & 3.0 & 1.681 & \underline{\textbf{0.521}} & 2.36 & \underline{\textbf{0.372}} \\
            & $\alpha$ & 6.4/7.8 & \underline{\textbf{1.6}} & 8.82/12.59 & 2.644 & \underline{\textbf{0.99}} & 2.058 \\
        \addlinespace
        zundel  & D & 2.4 & 0.4 & 0.371 & \underline{\textbf{0.049}} & 0.15 & \underline{\textbf{0.046}} \\
        & $\alpha$ & 3.8/0.97 & \underline{\textbf{0.1}} & 1.20/1.06 & \underline{\textbf{0.129}} & \underline{\textbf{0.09}} & \underline{\textbf{0.090}} \\
        \addlinespace
        liquid water    & D & - & 15 & \underline{\textbf{0.852}} & 1.260 & \underline{\textbf{0.7}} & 0.959 \\
            & $\alpha$ & 5.8/19 & \underline{\textbf{2.1}} & 16.28/20.38 & 5.898 & \underline{\textbf{0.48}} & 5.303 \\
    \bottomrule
\end{tabular}
\footnotetext{Results for all models except TACE are taken from refs.~\citep{HotPP, TNEP}. For polarizability, some models predict the $\ell=0$ and $\ell=2$ components individually, and their errors are given as values separated by a slash. In this benchmark, we report errors on the validation set in accordance with previous studies.}
\end{table}

\subsection{Liquid water: structure, dynamics, and spectra}
    We assess the performance of TACE on liquid water using a dataset of 1593 configurations labeled at the revPBE0-D3 level ~\citep{benchmark-h2o}, which is known to provide a reliable description of the structure and dynamics of water across a range of pressures and temperatures~\citep{h2o-quantum-effect-spectrum4}. Listed in Table~\ref{tab:h2o_results},  TACE attains the lowest RMSE of energy per atom and forces, comparing against invariant models (BPNN and DeepMD~\citep{BPNN, DeepMD}), linear ACE-style models, and recent equivariant message-passing models (NequIP, MACE and CAMP~\cite{NequIP, MACE, CAMP}). Notably, similar to ref.~\citep{MACE-application}, a small-scale TACE model with invariant messages ($\mathrm{L_{max}=0}$) already outperforms top-performing models, indicating that the irreducible Cartesian framework is highly data-efficient and does not require large, high-rank tensor channels to reach SOTA accuracy.
    
   We then test whether this accuracy translates into physically correct dynamics and performed three independent 300-ps simulations of bulk water at 300 K, 500 K, 1000 K, 1500 K, and 2000 K using the $\mathrm{L_{max}=0}$ model. The O-O radial distribution function (RDF) is shown in Fig.~\ref{fig:h2o}~(a). At 300 K, the calculated RDF agrees remarkably well with experimental results~\citep{benchmark-h2o-X-ray}. We also computed the mean square displacement (MSD) and obtained the diffusion coefficient using the Einstein relation, yielding $ 0.26(1), 1.58(1), 3.17(7), 4.66(7), 5.90(28)  \times 10^{-5}\ \mathrm{cm^2/s}$ at 300, 500, 1000, 1500, 2000K, which is in excellent consistency with the AIMD value of $ 2.67\times 10^{-5}\ \mathrm{cm^2/s}$ in 300K~\citep{benchmark-h2o-dft-diffusivity}. The TACE potential maintains stable MD simulations up to 2000 K, highlighting its robustness. 

    A central goal of this work is to explore whether TACE can simultaneously and accurately predict both scalar and tensorial properties. We therefore investigate whether TACE can describe dipole moments and polarizabilities of various water systems, and reproduce the infrared and Raman spectra of liquid water. For the dipole moment, we directly predict a vector $\nu = 1$. For the polarizability, since it is a traceless and fully symmetric tensor, we predict two components: the trace part ($\ell= 0$) and the traceless fully symmetric part ($\ell = 2$). The prediction accuracy is reported in terms of the relative RMSE~\citep{SA-GPR} in Table~\ref{tab:dipole_and_polar} and the explicit formula can be found in ref.~\citep{T-EANN}. TACE achieved very high accuracy in predicting dipole moments, achieving the best performance in three of the four systems considered, and shows competitive polarizability accuracy overall. However, some anomalous values can be observed in the Table~\ref{tab:dipole_and_polar}, despite its strong performance across other systems. We believe that this is due to the fact that the polarizability labels are not true symmetric tensors; instead, they are constructed from the instantaneous dielectric tensor using the tensorial extension of Clausius-Mossotti relation~\citep{CM-relation}. HotPP fits it as a reducible Cartesian tensor by predicting a generic rank-2 tensor, which is later symmetrized, and includes an additional scalar corresponding to the diagonal element. By contrast, our model enforces full tensor symmetry by explicitly summing the $\ell = 0$ and $\ell = 2$ components. Thus, the slightly higher polarizability error is therefore best understood as a limitation of dataset labels~\citep{SA-GPR} due to intrinsic noise, not of the representation itself. 
    
    To obtain the infrared and Raman spectra, we employed an NPT-equilibrated configuration of 512 water molecules from ref.~\citep{HotPP}. This was followed by an additional 100~ps equilibration in the NVT ensemble using TACE trained on the dataset of ref.~\citep{h2o-ir-raman-dataset}. Subsequently, three production simulations were carried out in the NVE ensemble, each with a duration of 200~ps. Throughout all simulations, a time step of 0.5~fs was applied. The infrared and Raman spectra were then obtained by performing Fourier transforms of the respective autocorrelation functions. From Fig.~\ref{tab:h2o_results}, we can see that our computed results are in good agreement with experiments. The Raman spectrum shows a slight blue shift in the high-frequency region, which can also be attributed to the non-strict symmetry of the polarizability labels. At the same time, our spectra were obtained without applying broadening. It should be emphasized that quantum effects in water can significantly influence both the RDF and the simulated spectra, as demonstrated in refs.~\cite{h2o-quantum-effect-spectrum1,h2o-quantum-effect-spectrum2,h2o-quantum-effect-spectrum3,h2o-quantum-effect-spectrum4}. Overall, our model achieves excellent performance in both stability and accuracy, while also demonstrating TACE's effective simulation of scalars, tensors, and spectra.

\begin{table}[h]
    \normalsize
    \centering
    \caption{RMSE for energy (E, meV/atom) and forces (F, meV/Å) for the GAP17 dataset. The best-performing results are both \textbf{bolded} and \underline{underlined}.}
    \label{tab:gap17_results}
    \begin{tabular}{cccccccccc}
        \toprule
            & GAP\citep{GAP} & MTP\citep{MTP} & REANN\citep{REANN} & NEP\citep{GPUMD} & NequIP\citep{NequIP} & HotPP\citep{HotPP} & TACE\\
        \midrule
        E & 46 & 35 & 31 & 42 & 17 & 16 & \underline{\textbf{8.55}} \\
        F & 1100 & 630 & 640 & 690 & 431 & 395 & \underline{\textbf{299.88}} \\
        \bottomrule
    \end{tabular}
    \footnotetext{Results for all models except TACE are from refs.~\citep{GPUMD, HotPP}.}
\end{table}

\begin{figure}[h]
    \centering
    \includegraphics[scale=0.4]{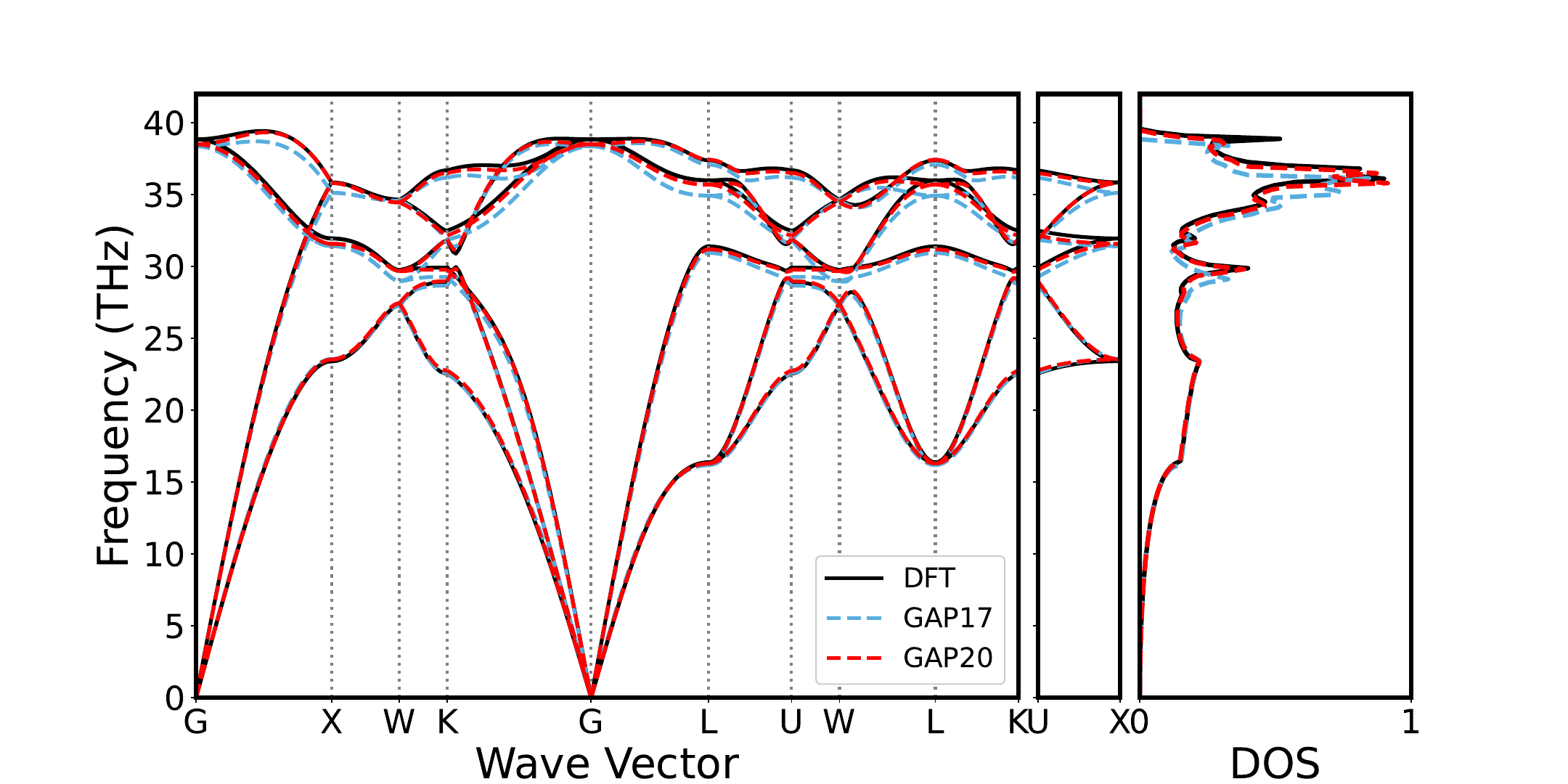}
    \caption{\textbf{Phonon Spectrum of Diamond.} Phonon dispersion relations of diamond predicted by TACE models trained on the GAP17~\citep{GAP17} and GAP20~\citep{GAP20} datasets, compared with DFT results~\citep{HotPP}.}
    \label{fig:gap17}
\end{figure}

\subsection{Hessians --- GAP17 and GAP20}

    Accurate prediction of higher-order derivatives by the MLIP is crucial for phonon behavior, as the curvature of the true PES around equilibrium points directly determines the phonon frequencies and dispersion. We benchmarked on the GAP17 dataset~\citep{GAP17} contains configurations representing a variety of carbon allotropes. On this dataset, TACE achieves an energy RMSE of 8.55 meV/atom and forces RMSE of 299.88 meV/Å, substantially outperforming other models in (Table~\ref{tab:gap17_results}). Accurate modeling of vibrational properties has historically posed challenges for carbon systems in ref.~\citep{GAP20}. To directly assess curvature quality, we compute phonon dispersions for diamond from the TACE potential, following the same protocol used in ref.~\citep{HotPP}, and Phonopy~\citep{Phonopy1,Phonopy2} to generate the phonon spectrum. As illustrated in Fig.~\ref{fig:gap17}, the phonon branches predicted by TACE agree well with DFT across the Brillouin zone. By contrast, we noticed that HotPP~\citep{HotPP} model trained on the GAP17 dataset, reported noticeable discrepancies in the high-frequency region near the $\Gamma$-point. To enable a more detailed comparison, we retrained our model using the more accurate and comprehensive GAP20 dataset~\citep{GAP20}, generated with the optB88-vdW. The resulting phonon spectrum is in near-perfect agreement with DFT calculations, confirming that TACE can reproduce both forces and second-derivative information across low fidelity and high fidelity DFT datasets.

 \begin{table}[h]
        \centering
        \caption{MAE for electric enthalpy (U, meV/atom), forces (F, meV/Å), stress (S, $\mathrm{meV/\text{\AA}^3}$), polarization (P, $\mathrm{me \cdot \text{\AA} / atom}$), polarizability ($\alpha$, $\mathrm{me \cdot \text{\AA}^2 \cdot V^{-1}}$) and Born effective charges (BEC, me) for BaTiO$_{3}$. For models trained with 59 configurations, the best-performing results are highlighted in both \textbf{bold} and \underline{underline}.}
        \label{tab:BaTiO3}
        \begin{tabular}{ccccc}
        \toprule
            & \multicolumn{3}{c}{$N_{\text{train}}=59$} & $N_{\text{train}}=75$ \\
            \cmidrule(lr){2-4} \cmidrule(lr){5-5}
            & FITACE & Allegro-pol-ZBL & FITACE-ZBL & Allegro-pol~\citep{Allegro-pol} \\
        \midrule
        U & 0.32 & 0.46 & \underline{\textbf{0.25}} & 0.5 \\
        F & \underline{\textbf{13.93}} & 27.6 & 14.10 & 14.6 \\
        S & 0.13 & 0.43 & \underline{\textbf{0.12}} & - \\
        P & 0.39 & 1.74 & \underline{\textbf{0.33}} & 1.4 \\
        $\alpha$ & 43.09 & 45.3 & \underline{\textbf{39.13}} & 15.2 \\
        BEC & \underline{\textbf{13.71}} & 21.2 & 13.73 & 14.1 \\
        \bottomrule
        \end{tabular}
        \footnotetext{
        Results for Allegro-pol trained with 75 configurations are taken from the official paper~\citep{Allegro-pol}, and it is unclear whether ZBL (Ziegler-Biersack-Littmark) potentials~\citep{ZBL} were used. All other models were trained by us using the publicly available 59 configurations, while Allegro-pol was trained using the official input files. For the 59-configuration models, we explicitly indicate whether ZBL was applied.
        }
\end{table}

\subsection{External field --- BaTiO$_{3}$}
    Modeling ferroelectric perovskites such as BaTiO$_{3}$ under external electric fields is challenging for first-principles methods. To assess whether TACE can learn such field-coupled responses, we use its field-embedded variant (FITACE), which incorporates the external electric field as an equivariant through the universal embedding framwork.

    We trained FITACE with the BaTiO$_{3}$ dataset introduced in ref.~\citep{Allegro-pol}, which consists of 59 frames extracted from MD simulations in the limit of zero electric field. We compare the unified model with the benchmark results (see Table~\ref{tab:BaTiO3}), and FITACE achieves greater accuracy due to its architecture and semi-local nature. This indicates that our framework is capable of faithfully reproducing various physical properties under external electric fields. When we additionally include a short-range ZBL term (FITACE-ZBL), the accuracy further improves and remains competitive with, or better than, Allegro-pol trained on a larger set of 75 configurations. These results indicate that the universal embedding framework makes TACE extendable, with minimal modification, to systems under external electric fields.

\subsection{Charged systems} 
    We evaluate charge-aware learning on the dataset constructed by Ko et al.~\citep{4G-HDNNPs}, which inculdes structures in various charge states where long-range electrostatics and charge transfer are decisive. Ag trimers with total charge of +1 and -1, ionic Na$_9$Cl$_8$ clusters formed by the removal of a neutral Na atom, carbon chains terminated with hydrogen atoms in both neutral and positively charged states, and a periodic system consisting of a gold cluster adsorbed on a MgO(001) slab. 
    
    Utilizing the universal embedding framework, energy and forces were benchmarked to illustrate the effects of including charges and total charge embeddings and the charge errors correspond to models trained jointly on energy, forces, and charge labels. All models, except ACE~\citep{ACE} and CACE-LR~\citep{LES2}, use LES or charge embeddings, either explicitly embedding charges or using charges as labels. Overall, the results are highly satisfactory, shown in Table~\ref{tab:charges}. One notable exception is the Ag systems, where our approach does not show a clear advantage in energy prediction. As long as embedding charges are used, incorporating the total charge as a continuous variable or omitting it has little effect on the error. We suspect this may be due to an inadequate embedding strategy for the total charge, a relatively low weight assigned to the energy term in the loss function, or possibly because competing models are overfitting, as suggested by the consistently low forces errors of our model.

\begin{table}[htbp]
    \centering
    \caption{RMSE for energy (E, meV/atom), forces (F, meV/Å) and charges (Q, me) for different charged systems. The best-performing results are highlighted in both \textbf{bold} and \underline{underline}.}
    \label{tab:charges}
    \begin{tabularx}{\textwidth}{ccccccccc}
    \toprule
    System &  & ACE\citep{ACE} & $\chi + \eta$(ACE)\citep{CCACE} & 3G-HDNNP\citep{3G-HDNNPs} & 4G-HDNNP\citep{4G-HDNNPs} & CACE-LR\citep{LES4} & TACE & \\
    \midrule
    C$_{10}$H$_2$/C$_{10}$H$_3^+$ & E & 0.76 & 0.75 & 2.045 & 1.194 & 0.73 & \underline{\textbf{0.21}} \\
    & F & 37.22 & 35.16 & 231.0 & 78.00 & 36.9 & \underline{\textbf{2.13}} \\
    & Q & - & 2.000 & - & 6.577 & - & \underline{\textbf{0.58}} \\

    Ag$_3^{+/-}$ & E & 809.62 & 0.21 & 320.2 & 1.323 & \underline{\textbf{0.162}} & 2.54 \\
    & F & 285.81 & 23.10 & 1913 & 31.69 & 29.0 & \underline{\textbf{12.22}} \\
    & Q & - & \underline{\textbf{0.415}} & - & 9.976 & - & 0.46 \\

    Na$_{8/9}$Cl$_8^+$ & E & 1.55 & 0.71 & 2.042 & 0.481 & \underline{\textbf{0.21}} & 0.55 \\
    & F & 41.72 & 12.35 & 76.67 & 32.78 & 9.78 & \underline{\textbf{1.06}} \\
    & Q & - & 7.020 & - & 15.830 & - & \underline{\textbf{1.40}} \\

    Au$_2$-MgO(001) & E & 2.56 & 1.63 & - & 0.219 & \underline{\textbf{0.073}} & 1.18 \\
    & F & 88.70 & 50.27 & - & 66.00 & 7.91 & \underline{\textbf{3.82}} \\
    & Q & - & 3.340 & - & 5.698 & - & \underline{\textbf{1.93}} \\
    \bottomrule
    \end{tabularx}
    \footnotetext{Results for all models except TACE are taken from refs.~\citep{LES4,4G-HDNNPs,CCACE}.}
    \end{table}

    \begin{table}[h]
    \tiny
    \centering
    \caption{Average distances of main chain carbon, carboxylic oxygen, and anhydride oxygen atoms from the Ag(111) surface (in \AA).}
    \label{tab:PTCDA}
    \begin{tabularx}{\textwidth}{ccccccccccc}
    \hline
    \toprule
    & & \multicolumn{5}{c}{\textbf{uMLIPs}} & &\textbf{finetune} & \multicolumn{2}{c}{\textbf{from-scratch}} \\
    \cmidrule(lr){4-8} \cmidrule(lr){9-9} \cmidrule(lr){10-11}
    & EXP & DFT & OrbV3 & 7Net & MACE &NequIP& Allegro & MACE & NequIP & TACE\\
    & \citep{PTCDA-exp} & \citep{PTCDA-dft} & con-inf & MF-omat & OMAT-medium & OAM-L & OAM-L &OMAT-medium& RECIO\citep{RECIO}& LES \\
    \midrule
    main chain carbon   & 2.86 & 2.97 & 3.38 & 3.22 & 3.21 & 3.73 &3.51 &3.00&2.98 & 2.97 \\
    carboxylic oxygen   & 2.66 & 2.67 & 3.07 & 3.03 & 3.40&3.29 &2.79&2.72 &2.68 & 2.69 \\
    anhydride oxygen  & 2.98 & 2.85 & 3.18 & 3.15 & 3.51&3.37&3.02 &2.99& 2.92 & 2.86 \\
    \midrule
    MAE	& - & -	&0.38& 0.30 & 0.54 & 0.63 & 0.28 & 0.07 & 0.03	& 0.01 \\
    \bottomrule
    \end{tabularx}
    \footnotetext{The benchmark result of the TACE model was obtained with a batch size of 16 at the 216th epoch and MAE are compared against DFT.}
    \end{table}

    \begin{table}[h]
    \centering
    \caption{Energy barriers (eV) for CO oxidation on various PdAg surfaces. Failures in searching TS are indicated by $\text{\texttimes}$.}
    \label{tab:CO}
    \begin{tabularx}{\textwidth}{ccccccccc}
    \toprule
    & & \multicolumn{4}{c}{\textbf{uMLIPs}} & \textbf{finetune} & \multicolumn{2}{c}{\textbf{from-scratch}} \\
    \cmidrule(lr){3-6} \cmidrule(lr){7-7} \cmidrule(lr){8-9}
    System & DFT & OrbV3 & MACE & NequIP & Allegro& MACE & NequIP & TACE \\
    & \citep{RECIO}  & con-inf & OMAT-medium & OAM-L & OAM-L & OMAT-medium &RECIO\citep{RECIO} & LES \\
    \midrule
    Pd100            & 0.63  & 0.60      & 0.66      & 0.65&0.88&0.69 & 0.67  & 0.65 \\
    Pd110            & 0.69  & 0.64      & 0.69      & 0.55&0.56&0.70 & 0.79  & 0.67 \\
    Pd111            & 1.39  & \texttimes & \texttimes &\texttimes &\texttimes&\texttimes & 1.35  & 1.29 \\
    Pd-TB            & 0.86  & 0.71      & 1.04 & 1.03&1.02& 1.17 & 0.91  & 1.12 \\
    Pd-cube          & 1.28  & 1.01      & 1.17 &1.17 &1.35& 1.23 & 1.18  & 1.23 \\
    Pd-S/Ag(111)     & 0.42  & 0.46 & \texttimes &\texttimes &\texttimes&\texttimes & 0.42  & 0.38 \\
    Pd-line/Ag(111)  & 0.55  & \texttimes & \texttimes &0.64&\texttimes& 0.51 & 0.58  & 0.51 \\
    Pd$_1$/Ag(111)      & 0.28  & 0.27  & \texttimes & 0.40 & 0.31 &\texttimes & 0.38  & 0.29 \\
    Pd$_4$/Ag(111)      & 0.23  & \texttimes & 0.28 & \texttimes & 0.32 & \texttimes & 0.21  & 0.20 \\
    Pd$_9$/Ag(111)      & 0.87  & \texttimes & \texttimes &\texttimes & \texttimes& \texttimes & 0.73  & 0.87 \\
    Pd$_{13}$/Ag(111)     & 0.74  & \texttimes & 0.36 &\texttimes& 0.60& 0.45  & 0.60   & 0.54 \\
    Pd$_{20}$/Ag(111)     & 0.90   & 0.69      & 0.89 &0.85&0.76& 0.92 & 0.99  & 0.91 \\

    \midrule
    MAE              &  -     & 0.11      & 0.11 & 0.10 & 0.15 & 0.11 & 0.07  & 0.07 \\
    \bottomrule
    \end{tabularx}
    \footnotetext{The benchmark result of the TACE model was obtained with a batch size of 16 at the 216th epoch and MAE are compared against DFT.}
    \end{table}

    \begin{table}[h]
    \centering
    \caption{Energy barriers (eV) for various Pd(111) and PdAg surfaces with different adsorbates. Failures in searching TS are indicated by $\text{\texttimes}$.}
    \label{tab:C2H2}
    \begin{tabularx}{\textwidth}{ccccccccc}
    \toprule
    & & \multicolumn{4}{c}{\textbf{uMLIPs}} & \textbf{finetune} & \multicolumn{2}{c}{\textbf{from-scratch}} \\
    \cmidrule(lr){3-6} \cmidrule(lr){7-7} \cmidrule(lr){8-9}
    System & DFT & OrbV3 & MACE & NequIP & Allegro& MACE & NequIP & TACE \\
    & \citep{RECIO}  & con-inf & OMAT & OAM-L & OAM-L & OMAT &RECIO\citep{RECIO} & LES \\
    \midrule
    Pd(111)-C$_2$H$_3$           & 0.87 & 0.72 & 0.74 &0.74&0.55& 0.78 & 0.73 & 0.79 \\
    Pd(111)-C$_2$H$_4$           & 0.79 & 0.70 & 0.76 &0.68 &0.68&0.72 & 0.74 & 0.71 \\
    Pd(111)-C$_2$H$_5$           & 0.79 & 0.67 & 0.61 & 0.63 &\texttimes&0.74 & 0.73 & 0.85 \\
    Pd(111)-C$_2$H$_6$           & 0.71 & \texttimes & 0.50 &0.44&0.54& 0.53 & 0.76 & 0.70 \\
    1 ML-Pd(111)-C$_2$H$_3$   & 0.67 & \texttimes & 0.72 &0.75 & \texttimes&\texttimes & 0.74 & 0.36 \\
    1 ML-Pd(111)-C$_2$H$_4$   & 0.69 & 0.65 & 0.65 &0.65 & 0.57&0.64 & 0.62 & 0.57 \\
    1 ML-Pd(111)-C$_2$H$_5$   & 0.90  & \texttimes & 0.60 &0.67& 0.59& 0.72 & 0.64 & 0.75 \\
    1 ML-Pd(111)-C$_2$H$_6$   & 0.45 & 0.45 & 0.39 &0.56 &0.39& 0.45 & 0.42 & 0.48\\
    0.25 ML-PdAg$_3$(111)-C$_2$H$_3$ & 0.58 & \texttimes & 0.61 &0.62&\texttimes& 0.60 & 0.51 & 0.63 \\
    0.25 ML-PdAg$_3$(111)-C$_2$H$_4$ & 0.37 & 0.59 & 0.52 &0.48& 0.51 & 0.47 & 0.44 & 0.43 \\
    0.25 ML-PdAg$_3$(111)-C$_2$H$_5$ & 0.42 & 0.67 & 0.61 & 0.66& 0.62 &0.68 & 0.71 & 0.62 \\
    0.25 ML-PdAg$_3$(111)-C$_2$H$_6$ & 0.17 & 0.33 & \texttimes&0.34& 0.28& 0.25 & 0.35 & 0.32\\
    1 ML-PdAg(111)-C$_2$H$_3$ & 0.58 & 0.52 & 0.52 & 0.53&0.43&0.52  & 0.48 & 0.49\\
    1 ML-PdAg(111)-C$_2$H$_4$ & 0.36 & \texttimes & 0.67 &0.42 &0.70&0.45 & 0.43 & 0.35 \\
    1 ML-PdAg(111)-C$_2$H$_5$ & 0.79 & 0.78 & 0.67 &0.66&0.65& 0.72 & 0.46 & 0.70\\
    1 ML-PdAg(111)-C$_2$H$_6$ & 0.36 & 0.44 & 0.41 &0.50&0.40&0.39 & 0.46 & 0.45 \\
    \midrule
    MAE & - & 0.11 & 0.13 & 0.13 & 0.17 &0.09 & 0.12 & 0.10 \\
    \bottomrule
    \end{tabularx}
    \footnotetext{The benchmark result of the TACE model was obtained with a batch size of 16 at the 216th epoch and MAE are compared against DFT.}
    \end{table}
    
\subsection{Advantages and Limitations of ACE --- REICO-AgPdCHO}\label{cart_vs_sph}

    Prior benchmarks use well-established databases designed to probe fundamental E/F/S potential energy surfaces or specific physical embeddings. These datasets most contains high-symmetry MD-derived structures which often clusters around near-equilibrium state, on which, as we shown, TACE achieves very high accuracy. Beside what we already shown, MLIP also also pursue generality, which is governed to a large extent by the chemical-space diversity of the training data. 
    Recent efforts such as RECIO~\citep{RECIO} sampling learns from randomized structures, reducing dependence on preselected configuration manifolds and pushing MLIPs toward DFT-like generality; likewise, recent new public datasets like OMat24\citep{OMat24}, also emphasizes non-equilibrium coverage. Yet while such non-standard or non-system-specific data broaden coverage, they also raise the bar for training. On the AgPdCHO dataset~\citep{RECIO}, for example, ACE-based models often require extensive hyperparameter tuning; otherwise training can exhibit a spontaneous RMSE rise and eventually fail (even when customized metric scaling or Huber loss is applied). By contrast, NequIP trained reliably across most settings, suggesting an architecture-data interaction rather than a purely data issue. We hypothesize that the randomly generated structures are hard to be simultaneously captured well lack enough physical constraints, which leads to conflicts with the strong inductive biases of ACE-based frameworks. In practice, we also observe that two-layer ACE-based models can, in some cases, become more sensitive to outliers.

    As a solution, we integrate LES as a plug-in long-range correction, which supplies appropriate physical constraints, that are especially helpful for models with short receptive fields, while remaining fully compatible with TACE's design. With LES add-on, TACE can effectively learn from the complex datasets generated by the RECIO method, overcoming the inherent limitation of ACE framework. We benchmark the TACE trained AgPdCHO EMLP(element-based machine learning potentials) with selected test cases from the original work that are representative in heterogeneous catalysis, including adsorption of PTCDA on Ag(111) (Table~\ref{tab:PTCDA}.), CO oxidation on various PdAg surfaces (Table~\ref{tab:CO}), and Acetylene hydrogenation on Pd(111), H covered Pd(111) and AgPd alloys (Table~\ref{tab:C2H2}), Baker reaction (see SI Table~S1). 
    We employed the Atomic Simulation Environment package~\citep{ASE} to perform the following structure optimizations and transition state (TS) searches. We have also included the latest uMLIP (MACE-OMAT-medium~\citep{MACE-uMLIP}, OrbV3-conservative-inf-omat~\citep{OrbV3}, 7Net-omat~\citep{7Net}, NequIP-OAM-L, Allegro-OAM-L~\citep{new-NequIP}) and fine-tuned MACE-OMAT-medium with REICO data for comparison. Overall, we find TACE delivers the most balanced, zero-shot performance among all models, sub-0.03 Å deviations for PTCDA/Ag(111) adsorption distances, MAE $\approx$ 0.07 eV for CO-oxidation barriers across diverse Pd/PdAg with 100\% TS success, and  within $\leq$ 0.1 eV of DFT across coverage/alloy conditions. We verified the reliability of the predicted TS geometries by confirming their structural correctness and ensuring that each valid TS exhibited exactly one imaginary frequency. For OrbV3-conservative-inf-omat, the model failed to obey the appropriate physical constraints. As a result, the models may encounter issues such as energy non-conservation in the NVE ensemble~\citep{NVE}, failure to locate transition state pathways~\citep{TS-fail}, or even extremely unusual and completely incorrect optimization trajectories (e.g., adsorption optimization of PTCDA using OrbV3-direct-inf; see SI Fig.~S2 ). It is also interesting to see the fine-tuned MACE-OMAT did not yield consistent gains and, in some cases, underperformed its pretrained counterpart. Our preliminary conclusion is that the conflict between ACE and RECIO persists, the randomized RECIO data contain a higher fraction of distributional outliers, which can confuse the model and degrade generalization.
    
\subsection{uMLIP --- TACE-MatPES}\label{tace-matpes}
    Here, we demonstrate the extrapolative capabilities of our model using a relatively small uMLIP dataset. In this case, we restrict the product basis with ``$\nu_1 \leq \nu_2$'' to further improve computational efficiency, and we train two types of models: an element-aware model and an element-agnostic model. Apart from the element-dependent part in the product basis, the rest of the architecture is kept identical. For consistency, we adopted the exact same data partitioning scheme as used in the MatPES publications~\citep{MatPES} of CHGNet~\citep{CHGNet}, M3GNet~\citep{M3GNet}, and TensorNet~\citep{TensorNet}. Using this setup, we evaluate the performance of the TACE-MatPES and compare its results on more comprehensive benchmarks. It can be seen that, whether using a small-parameter model or an extremely large-parameter model, we achieve superior results. Notably, TACE offers great flexibility in controlling model parameters; for model covering 89 elements, it allows us to freely choose between 1M and 30M parameters, and even smaller or larger parameter counts are permissible. For further comparisons regarding model size, hyperparameters, and speed, please refer to SI~2.1.

    \begin{table}[h]
    \centering
    \caption{MAE for energy per atom (E, meV), forces (F, meV/$\mathrm{\AA}$), stress (S, GPa), structural similarity (d), formation energy per atom ($\mathrm{E_f}$, meV/atom), bulk modulus ($\mathrm{K_{VRH}}$, GPa), shear modulus ($\mathrm{G_{VRH}}$, GPa) and constant-volume heat capacity ($\mathrm{C_v}$, $\mathrm{J \cdot mol^{-1} \cdot K^{-1}}$). Off-equilibrium forces ($\mathrm{\frac{F}{F_{DFT}}}$). Number of sarameters. (↑/↓) stands for higher/lower the better. The best-performing results are highlighted in both \textbf{bold} and \underline{underline}.}
    \label{tab:uMLIP}
    \begin{tabularx}{\textwidth}{rcccccc}
    \toprule
    & \multicolumn{6}{c}{\textbf{MatPES-PBE}} \\
    \cmidrule{2-7}
    & M3GNet~\citep{M3GNet} & CHGNet\footnotemark[1]~\citep{CHGNet} & TensorNet~\citep{TensorNet}  & TACE & TACE\footnotemark[3] & TACE\\
    & - & - & - & Element-Agnostic\footnotemark[2]~\citep{MACE} & ... & Element-Aware \\
    E ↓ & 45 & \underline{\textbf{31}} & 36 & 35& ... &\underline{\textbf{31}}\\
    F ↓ & 181 & 136 & 148 & 105 & ...&\underline{\textbf{101}}\\
    S ↓ & 0.888 & 0.642 & 0.700 & 0.655& ... &\underline{\textbf{0.607}}\\
    d ↓& 0.42 & 0.43 & \underline{\textbf{0.37}} & 0.39& ... & 0.38\\
    $\mathrm{E_f}$ ↓ & 110 & 82 & 81 & 80 &...& \underline{\textbf{72}} \\
    $\mathrm{K_{VRH}}$ ↓ & 26 & 24 & 18 & 17& ... & \underline{\textbf{16}} \\
    $\mathrm{G_{VRH}}$ ↓ & 25 & 21 & 15 & 15&... & \underline{\textbf{13}}\\
    $\mathrm{C_v}$ ↓ & 27 & 23 & 13 & 12&... &\underline{\textbf{11}}\\
    $\mathrm{\frac{F}{F_{DFT}}}$ ↑  & \underline{\textbf{0.97}} & 0.91 & 0.93 & 0.93 &...& 0.94 \\
    Parameters  & 0.66M & 2.70M & 0.84M & 0.98M&... & 29.8M\\
    \bottomrule
    \end{tabularx}
    \footnotetext[1]{CHGNet includes magmoms information. So there will be better performance in terms of energy.} 
    \footnotetext[2]{We follow MACE in the naming of model modules. ``Element-Agnostic" indicates that the parameters of this module (product basis here) do not depend on the elements. The results of MACE-MATPES-PBE-0 are not included here, as the model was pretrained on OMat24 and subsequently finetuned, with the MatPES dataset further subjected to an additional filtering step.}
    \footnotetext[3]{The ellipsis indicates that we can flexibly adjust the number of parameters by changing the size of the hidden channels and by choosing whether or not to include element dependence, ranging from 0.98M to 29.8M.}
    \footnotetext{We do not run MD benchmarks for this, as they can only be run by the Materials Virtual Lab.}
    \end{table}

\section{Discussion}\label{discussion}
    
    For the work presented in this paper, we believe that the effectiveness and accuracy of Cartesian models have already been well demonstrated. In our view, the most effective form of equivariance arises from tensor product operations (irreducible Cartesian tensor products/contractions, SO(3)/SO(2) spherical tensor products, and a series of other derived methods developed to accelerate computations). For atomic cluster expansions, the basis construction is typically not modified or augmented with additional nonlinear operations, as such changes may disrupt the encoded many-body structure and degrade the model accuracy. Introducing arbitrary nonlinearities at this stage can destroy the systematic body-order information learned by the model and therefore is generally avoided. For the spatial-domain formulation of ACE, however, we have verified its accuracy on the 3BPA benchmark. The results show that the spatial implementation achieves comparable accuracy to the frequency-domain formulation. Combined with the strong performance of ACE-based models on 3BPA, this observation further supports the validity and effectiveness of the spatial-domain ACE formulation. 

    For the Cartesian formulation, however, we also see current limitations in the implementation of irreducible Cartesian tensor products/contractions, and hereby discuss the future road-map. Rgarding the irreducible Cartesian tensor products/contractions, present implementation typically forms tensor products over all valid combinations ($\nu_1, \nu_2, \nu_o$, not path here) and then extracts the highest-weight irreducible component via ICTD. In contrast, a single path spherical tensor product can generate multiple irreducible components; importantly, Cartesian tensors can also yield multiple irreducible components from a single tensor product/contraction. The specific implementation approach has already been outlined in Fig.~\ref{fig:arch}a. Consider a rank-$\nu$ Cartesian tensor with weight $\mathrm{\nu-1}$. Such a tensor has $3^{\nu}$ components, but only $\mathrm{2\nu-1}$ of them are useful. In principle, one can project with ICTD, map to spherical irreps (retaining the $\mathrm{2\nu-1}$ nonzeros), then reassemble a rank-$\mathrm{(\nu-1)}$ Cartesian tensor of weight $\mathrm{(\nu-1)}$. However, this approach introduces certain issues in Cartesian space. For example, it becomes unavoidable to use a third tensor at the edge level for the tensor product, which leads to performance degradation due to the exponential growth of Cartesian tensors. How to improve this design to achieve better accuracy and efficiency remains an open question for future research. It is also known that SO(2) spherical tensor products significantly simplify computational cost, making $\mathrm{L_{max}}$ or  $\mathrm{\ell_{max}}$ = 6 feasible. In Cartesian space, it may be advantageous to use symmetric edge attributes. For example, if we rotate all edges to the direction $[1, 0, 0]$ (i.e., along the $x$-axis), then fully symmetric edge attributes with a trace will have only a single nonzero element. This implies that tensor products could effectively reduce to simple indexing and scaling operations. Meanwhile, since the Cartesian tensor product does not involve Clebsch-Gordan coefficients, the tensor product can be reduced to simple matrix multiplication. Therefore, its gradient is very simple. Hybrid designs are also viable, for example, applying Cartesian tensor products with spherical representations. 
    
\section{Conclusion}\label{conclusion}
    In this work, TACE establishes a systematic design paradigm for irreducible Cartesian equivariant models. In addition, TACE realizes atomic cluster expansion with coupling in frequency domain, while further extending the formulation to a spatial-domain ACE representation. Through an explicit basis transformation, the spatial-domain ACE can be naturally mapped into Cartesian space, thereby providing a unified bridge between spherical theory, spatial ACE formulations, and Cartesian tensor representations. TACE accommodates tensors of arbitrary rank and supports both reducible, irreducible Cartesian and spherical representations. The framework can consistently extract irreducible components of different weights, enabling flexible selection and manipulation of physically meaningful tensor contributions. Furthermore, the model is capable of directly predicting tensorial properties, and through a universal embedding mechanism, we can flexibly incorporate and control various factors (both invariant and equivariant) that influence the predicted physical quantities. Although we only demonstrate applications to external fields, charges, and fidelities in this work, the framework can be readily extended to a wide range of other tensorial observables. Taken together, these results demonstrate that TACE provides both theoretical consistency and practical flexibility, while maintaining competitive accuracy and efficiency. We therefore expect that frameworks of this type can serve as a general backbone for next-generation machine learning interatomic potentials, applicable across a broad range of problems in chemistry and physics. Given their generality, scalability, and ability to directly model complex tensorial properties, it is reasonable to anticipate that MLIPs based on our frameworks will become a primary computational engine for routine atomistic simulation and property prediction.

\section{Methods}\label{methods}

    \subsection*{Software}
    All experiments were conducted using the TACE software available at \url{https://github.com/xvzemin/tace} and Python version 3.12.11. The TACE-MatPES model was trained using PyTorch 2.8.0+cu128 and PyTorch Lightning 2.5.5, while all other results were obtained based on PyTorch 2.6.0+cu124 and PyTorch Lightning 2.5.2. For other models, we use orb-models 0.5.2, mace-torch 0.3.12(uMLIP), mace-torch 0.3.14(urea and speed test), cuequivariance 0.3.0, sevenn 0.11.1.dev2, allegro-pol 0.0.3 and nequip 0.15.0. In addition, most of the plotting relies on Matplotlib~\citep{matplotlib}.
    \subsection*{Bechmarks}
    3BPA: For the 3BPA dataset, we also utilize a 5~\AA{} cutoff~\citep{LinearACE}. The training set contains 500 structures, with 10\% reserved for validation. Testing is performed on four separate subsets corresponding to temperatures of 300~K, 600~K, 1200~K, and a dihedral scan, comprising 1,669; 2,138; 2,139; and 7,047 structures, respectively. 

    Water: For the liquid water dataset~\citep{benchmark-h2o}, a cutoff of 6~\AA{} is applied. Following the protocol used in CAMP~\citep{CAMP}, the dataset is split into 1,432 configurations for training, 79 for validation, and 81 for testing. The dipole moment ($\mu$) and polarizability ($\alpha$) for various water systems were taken from ref.~\citep{SA-GPR}. Values of $\mu$ and $\alpha$ for molecules such as $\mathrm{H_2O}$, $(\mathrm{H_2O})_2$, and $\mathrm{H_5O_2}^+$ were calculated using the coupled cluster singles and doubles (CCSD) method with the d-aug-cc-pVTZ basis set. For liquid water, the dataset was generated using the Perdew-Burke-Ernzerhof (PBE) exchange-correlation functional together with ultra-soft pseudopotentials (USPPs). A cutoff of 6~\AA\ was employed for training the dipole and polarizability models, except for $(\mathrm{H_2O})_2$, for which a cutoff of 10~\AA\ was used. The training and validation splits were either 500:500 or 700:300. The dataset employed for IR and Raman simulations comprises 1,888 structures calculated using the strongly constrained and appropriately normed (SCAN) functional. Our model attains a RMSE of 2.43meV/atom for energy, 30.97meV/\AA\ for forces, and 1.00 meV/atom for virials. The train-test split follows the TNEP scheme~\citep{TNEP}, with 1,510 structures used for training and 378 for testing, and an additional 10\% of the training set reserved for validation. The MD system simulated consists of 512 water molecules, corresponding to a density of 1 $\mathrm{g/cm^3}$. A time step of 1 fs was employed, and temperature control is maintained using a Nosé-Hoover thermostat. 

    GAP17 and GAP20: In the GAP17 dataset~\citep{GAP17}, a 5~\AA{} cutoff is used. The training subset includes 4,080 structures, with 10\% held out for validation purposes, while the test set contains 450 configurations. Similarly, the GAP20 dataset~\citep{GAP20} applies the same cutoff, with a total of 6088 structures in dataset. Of these, 10\% are used for validation and an additional 10\% for testing.

    BaTiO$_3$: The BaTiO$_3$ dataset was generated via active learning from MD simulations in the temperature range of 300-400 K. The original Allegro-pol~\citep{Allegro-pol} model was reported to be trained with 75 configurations. However, only 59 configurations are available in their public GitHub repository. Therefore, we retrained the model using these 59 configurations for a fair comparison. The training followed a protocol similar to the official setup, employing a cutoff radius of 6 Å, with 47 configurations used for training and 12 for validation. Model performance was evaluated on the validation set. The dataset was computed at the PBE level under a small external electric field of 0.36 MV·cm$^{-1}$, and the final reference data correspond to extrapolation to the zero-field limit.

    Charges: The datasets for different charge states and charge-transfer systems~\citep{4G-HDNNPs} were generated using the PBE functional with a light computational setting. Spin-polarized calculations were performed for the Au$_2$-MgO, NaCl, and Ag$_3$ systems. The Ag, NaCl, carbon-chain, and Au$_2$-MgO datasets contain 11,013, 5,000, 10,019, and 5,000 structures, respectively. For each dataset, 10\% of the configurations were used as the validation set, and all reported errors are evaluated on the validation set. For the charge embedding benchmarks of the Ag, NaCl, carbon-chain, and Au$_2$-MgO systems, the cutoff radii are 5.29, 5.29, 4.23, and 5.5~\AA, respectively. When using charges together with energy and forces as training labels, the corresponding cutoff values are 6.0, 6.0, 4.23, and 6.0~\AA, respectively.
    
    Urea: For this dataset, our model uses a fully unified set of parameters. Even in single-fidelity training, the model incorporates basis set labels, with a uniform cutoff of 5~\AA.

    PdAgCHO: PdAgCHO dataset was generated using the RECIO method~\citep{RECIO} and contains a total of 116,516 structures. For fine-tuning, 100,000 structures were randomly selected, while the remaining configurations were used for validation. For from-scratch training, we use all structures, and the learning rate scheduler adopts warmup followed by cosine annealing. All structures were computed at the PBE level. The fine-tuning models and TACE models was carried out by us with a uniform cutoff of 6 Å. It should be noted that the NequIP~\citep{NequIP} model was not trained by us, and that the dataset partitioning scheme employed in NequIP differs from ours.

    MatPES: The MatPES dataset~\citep{MatPES} was generated for uMLIPs, with calculations performed under both PBE and r2SCAN functionals. In this work, we only use the PBE subset and adopt the same data splitting scheme as in the original publication.

    \subsection*{Training details}
    All models are trained on a single RTX 4090 GPU, and if the memory is insufficient, a single A800 GPU is used instead. For all training, edge attributes were truncated at $\mathrm{l_{max}} = 3$, and the number of layers was set to $T = 2$. All models were trained with either the 64-0 or 64-2 (channel–$\mathrm{L_{max}}$) configuration, except for rMD17, which employed the 128-2 setting. Each combination was restricted to a single path, and the number of channels $c$ in both the atomic basis and the product basis was kept identical. The radial basis was constructed with 8 or 10 functions, which could be either fixed or trainable depending on the model setting. For the radial embedding, the MLP architecture was chosen as [64, 64, 64] or [64, 64, 64, 1024]. The SiLU activation function was consistently applied to all nonlinear operations. All models were optimized using Adam~\citep{Adam} with the AMSGrad variant~\citep{AMSgrad}, together with the ReduceLROnPlateau learning-rate scheduler implemented in PyTorch~\citep{Pytorch} and pytorch-cosine-annealing-with-warmup for PdAgCHO. The loss function is defined as the weighted sum of the mean squared errors (MSE) for energy per atom, forces, virials, with task-specific weighting hyperparameters and huber loss for TACE-MatPES. For more detailed hyperparameters see SI Table S5.

\section*{Data availability}
    Although multiple random seeds were not used in this study for efficiency, most tests were performed with seed = 42. We ensure that all results are fully reproducible. All datasets will be made available on GitHub \url{https://github.com/xvzemin/tace-fit}.

\section*{Code availability}
    The implementation of TACE is available on GitHub \url{https://github.com/xvzemin/tace}.

\section*{Acknowledgements}
    The authors thank Bin Jiang, Junfan Xia, Chenyu Wu, Changxi Yang, Junjie Wang and Shihao Shao for fruitful discussions. This work was supported by the National Natural Science Foundation of China NSFC (22433004 and 22403064) and ShanghaiTech University.

\section*{Author contribution}
    P.H. and W.X. conceived the project and guided the research the project. The TACE code, equations, figures, tables and benchmark results were prepared by Z.X. All authors edited and revised the manuscript.

\section*{Competing interests}
    The authors declare no competing interests.
    
\backmatter

\newpage
\bibliography{references}
\end{document}